\documentclass{article}

\usepackage{microtype}
\usepackage{graphicx}
\usepackage{subfigure}
\usepackage{booktabs} 

\usepackage{hyperref}

\usepackage[accepted]{icml2025}

\usepackage{amsmath}
\usepackage{amssymb}
\usepackage{mathtools}
\usepackage{amsthm}

\usepackage[capitalize,noabbrev]{cleveref}

\theoremstyle{plain}
\newtheorem{theorem}{Theorem}[section]

\theoremstyle{definition}

\theoremstyle{remark}

\usepackage[textsize=tiny]{todonotes}

\usepackage{bm}
\usepackage{graphicx}
\usepackage{subcaption}
\usepackage{tabularx}
\usepackage{booktabs}
\usepackage{multirow}
\DeclareUnicodeCharacter{2212}{\textminus}
\usepackage{wrapfig}

\icmltitlerunning{Behavior Cloning with Information Bottleneck}

\begin{document}

\twocolumn[
\icmltitle{Rethinking Latent Redundancy in Behavior Cloning: An Information Bottleneck Approach for Robot Manipulation}




\begin{icmlauthorlist}
\icmlauthor{Shuanghao Bai}{1}
\icmlauthor{Wanqi Zhou}{1}
\icmlauthor{Pengxiang Ding}{2,3}
\icmlauthor{Wei Zhao}{2}
\icmlauthor{Donglin Wang}{2}
\icmlauthor{Badong Chen}{1}
\end{icmlauthorlist}

\icmlaffiliation{1}{Institute of Artificial Intelligence and Robotics, Xi’an Jiaotong University, China.}
\icmlaffiliation{2}{Westlake University}
\icmlaffiliation{3}{Zhejiang University}

\icmlcorrespondingauthor{Badong Chen}{chenbd@mail.xjtu.edu.cn}
\icmlcorrespondingauthor{Donglin Wang}{wangdonglin@westlake.edu.cn}

\icmlkeywords{Behavior cloning, Information bottleneck, Robot manipulation}

\vskip 0.3in
]



\printAffiliationsAndNotice{}  

\begin{abstract}
Behavior Cloning (BC) is a widely adopted visual imitation learning method in robot manipulation.
Current BC approaches often enhance generalization by leveraging large datasets and incorporating additional visual and textual modalities to capture more diverse information.
However, these methods overlook whether the learned representations contain redundant information and lack a solid theoretical foundation to guide the learning process.
To address these limitations, we adopt an information-theoretic perspective and introduce mutual information to quantify and mitigate redundancy in latent representations.
Building on this, we incorporate the Information Bottleneck (IB) principle into BC, which extends the idea of reducing redundancy by providing a structured framework for compressing irrelevant information while preserving task-relevant features.
This work presents the first comprehensive study on redundancy in latent representations across various methods, backbones, and experimental settings, while extending the generalizability of the IB to BC.
Extensive experiments and analyses on the CortexBench and LIBERO benchmarks show consistent performance improvements with IB across various settings, underscoring the importance of reducing input data redundancy and highlighting its practical value for real-world applications.
Project Page: \href{https://baishuanghao.github.io/BC-IB.github.io}{BC-IB Website}.
\end{abstract}

\vspace{-20pt}
\section{Introduction}
\label{sec:intro}
\vspace{-2pt}

\begin{figure}[ht]
\begin{center}
\centerline{\includegraphics[width=\columnwidth]{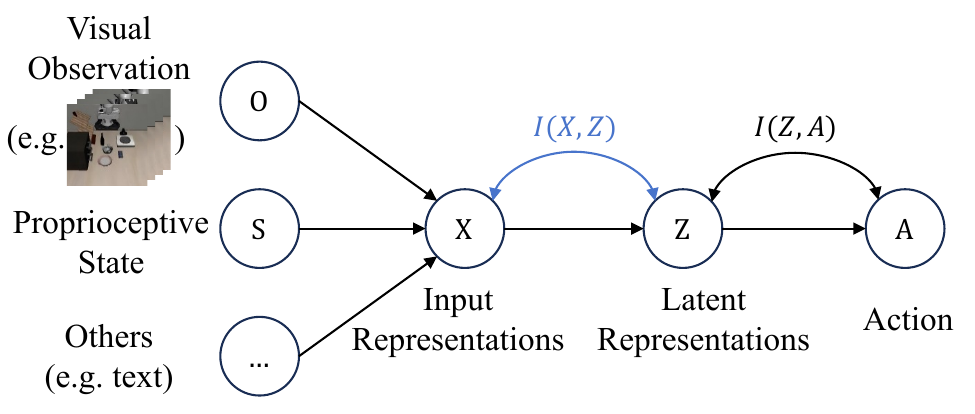}}
\caption{
Policy architecture of BC. Current BC methods (black arrows) do not impose restrictions on the latent representations $Z$, potentially allowing redundant information from the input representations $X$.
}
\label{fig:intro}
\end{center}
\vskip -0.25in
\end{figure}

Behavior Cloning (BC), one of the simplest and most widely used methods in Imitation Learning (IL), learns a mapping from states to actions by training on state-action pairs from expert demonstrations. 
BC has been widely studied in autonomous driving~\cite{bain1995framework, torabi2018behavioral}, robotics control~\cite{argall2009survey} and game AI~\cite{pearce2022counter}. 
In robot manipulation, BC has become a foundational approach, enabling robots to replicate expert actions based on sensory inputs such as images or proprioception information like gripper states. 
To enhance the generalization of robots, most BC methods focus on incorporating large datasets of human or manipulation videos~\cite{jang2022bc, zhang2024hirt, cheang2024gr, saxena2025matters, cui2025openhelix}, or integrating additional text and visual information~\cite{jia2024chain, wen2023any, hu2024video, zhang2025up}.
While these methods have made significant progress in improving generalization by leveraging more diverse information, they often neglect a critical aspect: whether the learned representations contain significant redundant information.

\textit{\textbf{Why do we need to explore this?}}
Firstly, the inherent challenges of input data redundancy remain largely unexplored in BC for robot manipulation, despite their potential impact on policy performance and generalization.
Secondly, most existing methods lack a solid theoretical foundation to guide the learning process.
This raises a key question: how can we formally characterize and reduce redundancy in inputs or representations in a theoretically grounded way?

\textit{\textbf{How to explore this?}}
As illustrated in~\cref{fig:intro}, in BC, the inputs are typically encoded into individual representations and concatenated to form the input representation $X$. This is then processed through a feature fusion module to produce the latent representation $Z$, which is subsequently decoded to predict the action $A$. The policy is optimized by minimizing the discrepancy between the predicted actions and the expert-provided actions.
In information theory, mutual information between $X$ and $Z$, denoted as $I(X, Z)$, measures the amount of information gained about one random variable by knowing the other. In BC, if output $Y$ can be well predicted by $Z$, reducing $I(X, Z)$ means continuously eliminating redundant information from $X$.

Taking a step further, an information-theoretic approach that balances the trade-off between representation complexity and predictive power offers a \textit{natural framework} to address the problem of latent representation redundancy and the lack of a solid theoretical foundation, namely information bottleneck (IB) principle~\cite{tishby2000information}. 
IB regularizes the representation $Z$ by minimizing the mutual information $I(X, Z)$ between $X$ and $Z$, while maximizing the mutual information $I(Z, A)$ between $Y$ and $A$. 
The first term $I(X, Z)$ represents the compression of the representation, where a smaller mutual information indicates a greater degree of compression and redundancy reduction, while $I(Z, A)$ ensures predictive power is maintained.   

Motivated by this information-theoretic approach, we make the first attempt in this work to study the impact of latent representation redundancy in BC for robot manipulation and extend the IB method to this context, where redundancy in latent representations is quantified by $I(X, Z)$. 
We conduct extensive experiments in various settings and analyses to validate its effectiveness, highlighting the benefits of reducing redundancy to enhance generalization in robotic tasks.
Additionally, we provide detailed theoretical analyses, including generalization error bounds, to validate its effectiveness.

\textbf{\textit{How to apply IB to the BC architectures, and what are its potential applications?}}
To ensure the generality of our findings, we categorize BC architectures based on their feature fusion methods into two types: spatial fusion and temporal fusion. This allows us to identify the applicable scenarios for each fusion method, and by incorporating IB, we uncover a series of interesting findings. 
Furthermore, our experiments reveal that regardless of the pre-training stage, the final fine-tuning phase, or the size of the dataset, incorporating IB by reducing redundancy enables the model to learn more robust features and improve performance, suggesting its potential applicability in these scenarios.

Our contributions are three-fold. 
(1) We extend the IB to BC and provide a comprehensive study on the impact of latent representation redundancy in BC for robot manipulation.
(2) We empirically demonstrate that minimizing redundancy in latent representations helps existing BC algorithms significantly improve generalization performance on the Cortexbench and LIBERO benchmarks across various settings, indirectly highlighting the considerable redundancy present in current robot trajectory datasets.
(3) We provide a detailed theoretical analysis explaining why IB enhances the transferability of BC methods.

\vspace{-2.5pt}
\section{Related Work}
\label{sec:rw}
\vspace{-2pt}

\textbf{Behavior Cloning in Robot Manipulation.}
Behavior Cloning (BC), first introduced by~\cite{pomerleau1991efficient}, is a well-known Imitation Learning (IL) algorithm that learns a policy by directly minimizing the discrepancy between the agent's actions and those of the expert in the demonstration data. 
To learn more generalizable representations, one class of visual representation learning methods pre-trains on large video datasets of robotics or humans, enabling rapid application of the pre-trained encoder to downstream robotic tasks. Notable examples include VC-1~\cite{majumdar2023we}, R3M~\cite{nair2023r3m}, and Voltron~\cite{karamcheti2023language} . 
Meanwhile, another line of research focuses on training on even more extensive and diverse datasets with larger models, such as Internet-scale visual question answering and robot trajectory data~\cite{brohan2023rt}, as well as a vast collection of Internet videos~\cite{cheang2024gr}.
Additionally, some methods further enhance generalization by incorporating additional sources of information. These include inferring textual descriptions based on the robot's current state~\cite{zawalski2024robotic}, leveraging visual trajectories~\cite{wen2023any} and generated images~\cite{tian2025predictive, hu2025video, zhang2025gevrm}, and integrating 3D visual information~\cite{goyal2023rvt}.
However, these methods have not deeply analyzed the redundancy in learned latent representations, and most also lack a solid theoretical foundation. 
Thus we extend the Information Bottleneck (IB) principle to BC, addressing this fundamental gap.

\textbf{Information Bottleneck in Robotics.}
The Information Bottleneck (IB) principle was first proposed in~\cite{tishby2000information} within the context of information theory. 
Since then, it has been widely applied in deep learning and various downstream tasks to balance the trade-off between representation accuracy and complexity, including classification~\cite{federici2019learning}, segmentation~\cite{bardera2009image, lee2021reducing}, and generative tasks~\cite{jeon2021ib}.
In robotics learning, IB has found notable applications in reinforcement learning, where some works maximize the mutual information between the representation and the dynamics or value function, while restricting the information to encourage the encoder to extract only task-relevant features~\cite{kim2019curiosity, bai2021dynamic, he2024bridging}. In imitation learning, it has been applied to alleviate the copycat problem from observation histories~\cite {wen2020fighting}.
Different from prior works, we introduce IB into Behavior Cloning to explore and empirically validate the redundancy in latent representations in robotics. 
Additionally, we demonstrate its effectiveness through detailed theoretical analyses.

\begin{figure*}[ht]
\begin{center}
\centerline{\includegraphics[width=0.95\textwidth]{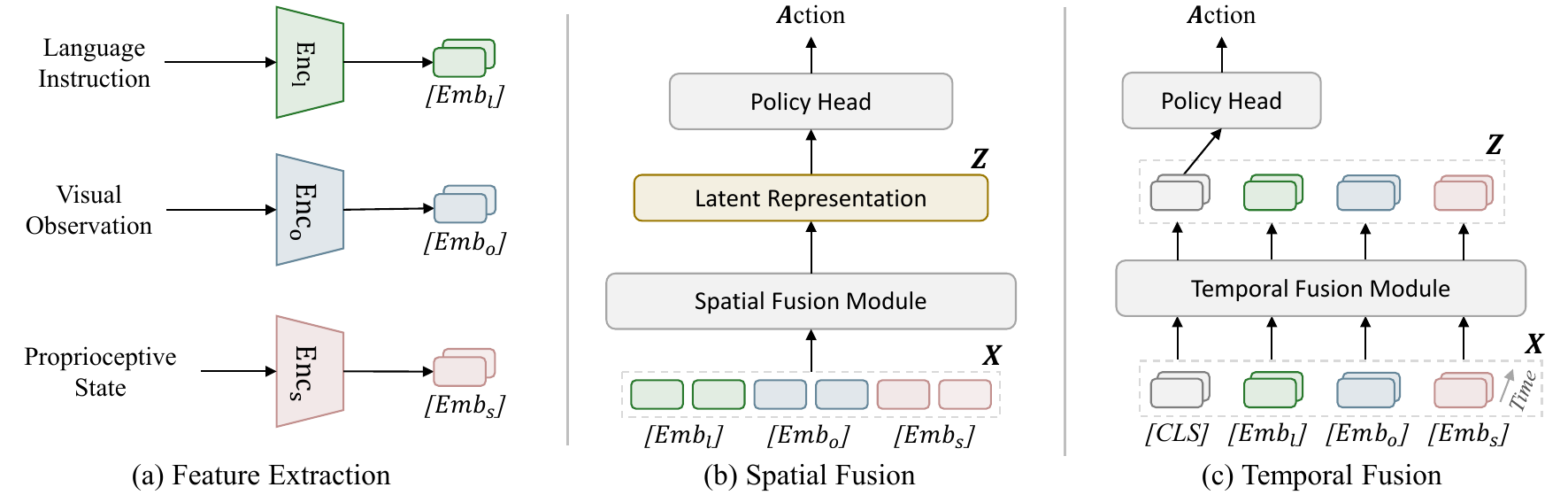}}
\caption{
Model architectures used in this study. Based on feature fusion methods, we categorize the BC methods in robot manipulation into two types: spatial fusion and temporal fusion. 
After extracting features from each modality a), spatial fusion b) extracts spatial features at a given time step or concatenates features across multiple time steps using encoders like MLPs or CNNs. Temporal Fusion c) fuses input features by modeling dynamic relationships and dependencies between time steps using RNNs or Temporal Transformers. 
The latent representations are then decoded into actions via the policy head.
}
\label{fig:model}
\end{center}
\vskip -0.25in
\end{figure*}

\section{Preliminary}
\label{sec:pre}

\subsection{Problem Setting of Behavior Cloning}
\label{subsec:ps}

BC can be formulated as the Markov Decision Process (MDP) framework~\cite{torabi2018behavioral}, which is often defined without an explicitly specified reward function, to model sequential action generation problems. The concept of rewards is replaced with supervised learning, and the agent learns by mimicking expert actions.
Formally, in robot manipulation, the state at each timestep consists of visual observations $o_t$, the robot’s proprioceptive state $s_t$, and optionally a language instruction $l$.
Let $x_t = (o_t, s_t, l)$ represent the overall state. 
The policy $\pi$ maps a sequence of states to an action: 
$\hat{a}_t = \pi(x_{t-\tau:t})$, where $\tau$ indicates the length of the state history. For simplicity, we set $\tau=1$.
The optimization process can be formulated as:
\begin{equation}\begin{aligned}\label{func:vanilla_bc}
\pi^* = \operatorname{argmin}_\pi \mathbb{E}_{(x_{t}, a_t) \sim \mathcal{D}_e}\left[\mathcal{L}\left(\pi\left(x_{t}\right), a_t\right)\right],
\end{aligned}\end{equation}
where $\mathcal{D}_e$ is expert trajectory dataset and $a_t$ is action labels. In vanilla BC, $\mathcal{L}$ typically represents the mean squared error (MSE) loss function for continuous action spaces, or cross-entropy (CE) loss for discrete action spaces. 
In this study, we adopt the continuous action spaces with MSE loss:
\begin{equation}\begin{aligned}\label{func:vanilla_bc_loss}
\mathcal{L}_{\mathrm{BC}}=\mathbb{E}_{\left(x_t, a_t\right) \sim \mathcal{D}_e}\left[\left\|\pi\left(x_t\right)-a_t\right\|^2\right].
\end{aligned}\end{equation}
Building on this vanilla BC loss, some methods also introduce alignment loss~\cite{jang2022bc, ma2024contrastive} and reconstruction loss~\cite{radosavovic2023real, karamcheti2023language}. However, in this study, to more clearly illustrate the relationship with representation redundancy, we focus solely on the vanilla BC loss.

\subsection{Mutual Information Neural Estimation}

Estimating mutual information between variables directly is challenging, thus we use Mutual Information Neural Estimation (MINE)~\cite{belghazi2018mutual} to estimate it. 
MINE is based on neural networks, which can efficiently handle high-dimensional, continuous, discrete, and hybrid data types without requiring assumptions about the underlying distributions.
MINE estimates mutual information by training a classifier to differentiate between samples from the joint distribution $P_{XZ}$ and the product of the marginal distributions $P_{X} \otimes P_{Z}$ of the random variables $X$ and $Z$.
MINE uses a lower bound for mutual information based on the Donsker-Varadhan representation~\cite{donsker1983asymptotic} of the Kullback-Leibler (KL) divergence:
\begin{equation}\begin{aligned}\label{func:mine}
\mathcal{I}(X&; Z):=\mathcal{D}_{KL}(P_{XZ} || P_{X} \otimes P_{Z}) \geq \widehat{\mathcal{I}}_\theta^{(DV)}(X; Z) \\
&:=\mathbb{E}_{P_{XZ}}\left[T_\theta(x, z)\right]-\log \mathbb{E}_{P_{X} \otimes P_{Z}}\left[e^{T_\theta(x, z)}\right],
\end{aligned}\end{equation}
where $T_\theta: \mathcal{X} \times \mathcal{Z} \to \mathcal{R}$ is a discriminator function modeled by a neural network with parameters $\theta$. 
We empirically sample from $P_{XZ}$, and for $P_{X} \otimes P_{Z}$, we shuffle the samples from the joint distribution along the batch axis.

\section{Pipeline of BC with IB}

\subsection{Model Architecture}
\label{subsec:ma}

Before introducing IB, we first define its input and latent representations.
Traditional IB methods~\cite{amjad2019learning, pacelli2020learning, wan2021multi} typically apply the bottleneck to a single modality (e.g., images or states) and their corresponding latent features, following the information flow $O \rightarrow Z \rightarrow A$.
In contrast, BC for robot manipulation is more complex than earlier control or single-modal tasks, as it requires models to process diverse, multimodal data. This data not only includes RGB images but may also incorporate the robot’s proprioceptive state, language instructions, and other modalities, making effective feature fusion essential. Strictly adhering to the conventional IB paradigm would involve constraining each input modality and its corresponding features separately, resulting in a pipeline that is inelegant, difficult to scale, overly complex, and unable to capture cross-modal associations.
Furthermore, previous work has shown that proprioceptive states can lead to overfitting~\cite{wang2024scaling}.

As a result, we do not treat image or other modalities separately as inputs to IB, as done in previous studies.
Instead, we concatenate features extracted from all modalities through respective feature extractors as our input $X$, i.e., 
\begin{equation}\begin{aligned}\label{func:feature}
x_t=\text{concat}(\text{Enc}_\text{o}(o_t), \text{Enc}_\text{s}(s_t), \text{Enc}_\text{l}(l)),
\end{aligned}\end{equation}
where $\text{Enc}_{(\cdot)}$ denotes the feature extractor of each modality.
This results in the information flow $O \rightarrow X \rightarrow Z \rightarrow A$, and brings several advantages:
(1) It enables unified redundancy reduction across all modalities.
(2) In practice, encoders are often frozen, and this approach is more effective under such conditions.
(3) It scales better to a broader range of robotic algorithms.
Then, regarding how to process the input $X$, or how to fuse information from multiple modalities into latent representations $Z$, we categorize BC methods in robot manipulation into two types based on their feature fusion strategies: spatial fusion and temporal fusion.

As illustrated in Figure~\ref{fig:model} (b), spatial fusion involves extracting spatial features from data at a given time step or concatenating features across multiple time steps along the feature dimensions. This approach does not explicitly differentiate between time steps but instead processes the aggregated features as a whole, emphasizing the modeling of inter-feature relationships. The spatial fusion module can be implemented using Multi-Layer Perceptrons (MLPs), Convolutional Neural Networks (CNNs), Spatial Transformers, or even simple concatenation operations.
These methods are primarily designed to learn highly generalizable visual encoders by leveraging large-scale human video datasets. The pretrained encoders are then fine-tuned for downstream robotic tasks~\cite{majumdar2023we, zeng2024learning}.

On the other hand, as illustrated in Figure~\ref{fig:model} (c), temporal fusion integrates input features by capturing dynamic relationships and dependencies across time steps. This enables the modeling of both long-term and short-term temporal dynamics in sequential data. Temporal fusion modules can be implemented using Recurrent Neural Networks (RNNs), Long Short-Term Memory networks (LSTMs), or Temporal Transformers.
These methods are commonly incorporated into approaches that utilize Transformer-based backbones~\cite{wu2023unleashing, li2023vision, liu2024libero}.

The latent representation $Z$, which integrates both spatial and temporal information, is then passed through a policy head to generate actions. 
Policy heads typically include MLPs, GMMs, diffusion policy~\cite{chi2023diffusion, reuss2024multimodal}, and other designs~\cite{gong2024carp}. We choose an MLP for clarity and ease of empirical analysis.

\subsection{Behavior Cloning with Information Bottleneck}

The Information Bottleneck (IB) principle is an information-theoretic approach aimed at extracting the most relevant information from an input variable $X$ with respect to an output variable, \textit{i.e.}, action $A$.
The central idea is to find a compressed representation $Z$ of $X$ that retains the relevant information needed to predict $A$, while discarding irrelevant parts of $X$ that do not contribute to predicting $A$. 
The relevant information is quantified as the mutual information $I(X; A)$, and the optimal representation $Z$ is the minimal sufficient statistic of $X$ with respect to $A$. 
In practice, this can be achieved by minimizing a Lagrangian that balances the trade-off between retaining predictive information and compressing the input, which can be formulated as:
\begin{equation}\begin{aligned}\label{func:ib}
\mathcal{L} = \beta I(X; Z) - I(Z; A),
\end{aligned}\end{equation}
where $\beta$ is the Lagrange multiplier that balances the trade-off between the compression ability and the predictive power. Thus~\cref{func:vanilla_bc_loss} can be modified as:
\begin{equation}\begin{aligned}\label{func:bcib}
\mathcal{L}_{\mathrm{BC-IB}}=\mathbb{E}_{(x_t, a_t) \sim \mathcal{D}_e}\left[\beta I(x_t, z_t) + \|\pi(x_t)-a_t\|^2\right],
\end{aligned}\end{equation}
where $z_t=F(x_t)$ and $F(\cdot)$ denotes the fusion module. 
When $\beta < 0$, the information from $X$ to $Z$ increases; when $\beta = 0$, the method reduces to vanilla BC; and when $\beta > 0$, the information from $X$ to $Z$ decreases.

\subsection{Theoretical Analysis}

We provide a theoretical analysis of our BC-IB objective in~\cref{func:ib}. 
We adapt \cref{theorem:1} and \cref{theorem:2} to reveal that the generalization error is upper-bounded by the mutual information between the input $O$ and the latent representation $Z$, following the information flow $O \rightarrow Z \rightarrow A$. Minimizing this mutual information tightens the bound and improves generalization.
However, when $O$ is diverse and multimodal, directly minimizing the mutual information between each modality and its corresponding $Z$ is computationally intractable and unnecessarily complex. To address this, we extract and concatenate features from all modalities into an intermediate feature $X$, obtain $Z$ via a fusion network $f$, and minimize the mutual information between $X$ and $Z$ instead.
To validate the compatibility of this paradigm with the original theorems, we present~\cref{theorem:ours}. The theorem establishes that, even if we optimize $I(X; Z)$ by applying the bottleneck at an intermediate feature level $X$, as long as $X$ preserves the essential structure of the original input $O$, we are effectively controlling $I(O;Z)$, with the difference bounded by a small constant $\delta$.

\begin{theorem}
\label{theorem:1}
Generalization Bound Adapted from~\cite{IB1}. Let \( S = \{(x_t, a_t)\}_{t=1}^n \) denote the training data sampled from the same distribution as the random variable pair \( (X, A) \). Given the policy \( \pi \) trained on \( S \), the generalization error is given by:
\begin{equation}
\Delta(S) = \mathbb{E}_{X, A}[\ell(\pi(X), A)] - \frac{1}{n} \sum_{t=1}^n \ell(\pi(x_t), a_t).
\end{equation}
Using the Probably Approximately Correct (PAC) bound framework and the Asymptotic Equipartition Property (AEP)~\cite{AEP}, with probability at least \( 1 - \delta \), the following upper bound on the generalization error holds:
\begin{equation}
\Delta(S) \leq \sqrt{\frac{2 I(X; Z) + \log \frac{2}{\delta}}{2n}},
\end{equation}
where \( I(X; Z) \) represents the mutual information between the input \( X \) and the intermediate representation \( Z \), and \( \delta \) is the confidence level.
Details of proof can be seen in Appendix A of \cite{IB1}.
\end{theorem}

\begin{theorem}
\label{theorem:2}
Generalization Bound Adapted from \cite{IB2}.
Let \( S = \{(x_t, a_t)\}_{t=1}^n \) denote the training data sampled from the same distribution as the random variable pair \( (X, A) \). The generalization error is approximately bounded by:
\begin{equation}\label{eq:theorem2}
\Delta(S) \propto \sqrt{\frac{I(X; Z \mid A) + I(\phi^S; S)}{n}}, 
\end{equation}
where \( \phi^S \) is the encoder mapping the input \( X \) to the intermediate representation \( Z \).
This bound indicates that the generalization error is:
\begin{itemize}
    \item Positively correlated with \( I(X; Z \mid A) \), which captures mutual information between the input \( X \) and the latent representation \( Z \), conditioned on the actions \( A \).
    This term reflects that the IB compresses \( X \) into \( Z \) while preserving the relevant information for predicting \( A \).
    \item Positively correlated with \( I(\phi^S; S) \), which reflects the information content of the representation \( \phi \) for the given dataset \( S \).
\end{itemize}
\end{theorem}

\begin{theorem}
\label{theorem:ours}
Optimization Gap under Different Input Compression.
Let \( o \to x \to z \) form a Markov chain, where \( o \) is transformed into \( x \) by a network \( f \), and \( x \) is further transformed into \( z \) by a network \( \phi \). 
Let \( \phi_o = f \circ \phi \). Define two optimization problems:
\begin{align}
    (\theta^\varepsilon, \phi_o^\varepsilon) & = \arg\min_{\theta, \phi_o} \mathbb{E}_{P_{\phi_o}(o,x,z)} \left[ \log \frac{P_{\phi}(z|x)}{P_{\phi}(z)} - \frac{1}{\beta} J(z; \theta) \right],  \\
    (\theta^\star, \phi_o^\star) & = \arg\min_{\theta, \phi_o} \mathbb{E}_{P_{\phi_o}(o,z)} \left[ \log \frac{P_{\phi_o}(z|o)}{P_{\phi_o}(z)} - \frac{1}{\beta} J(z; \theta) \right]. 
\end{align}
Let $J^\varepsilon = \mathbb{E}_{P_{f^\varepsilon, \phi^\varepsilon}(o, x, z)}[J(z; \theta^\varepsilon)], \quad 
J^\star = \mathbb{E}_{P_{\phi_o^\star}(o, z)}[J(z; \theta^\star)]$.
Assume the mutual information gap satisfies the following condition: for any $\delta$, we have

\begin{equation}
I(o, z; \phi_o^\varepsilon) - I(o,z;\phi_o^*) \leq \frac{\delta}{\beta}.
\end{equation}

Then, the gap between the two optimizations is bounded as:
\begin{equation}
|J^\star - J^\varepsilon| \leq \delta.
\end{equation}
The detailed proof can be found in Appendix \ref{proof:ours}.
\end{theorem}

\section{Experiments}

\subsection{Embodied Evaluation}

\textbf{Simulation Benchmarks.} 
We mainly evaluate BC with IB across two benchmarks, CortexBench~\cite{majumdar2023we} and LIBERO~\cite{liu2024libero}. 
CortexBench is a single-task benchmark. For validation, we selected four imitation learning-related simulators, encompassing a total of 14 tasks: Adroit (2 tasks)~\cite{rajeswaran2018learning}, Meta-World (5 tasks)~\cite{yu2020meta}, DMControl (5 tasks)~\cite{tassa2018deepmind}, and TriFinger (2 tasks)~\cite{wuthrich2021trifinger}.
During evaluation, the number of validation trajectories is set to 25, 10, 25, and 25, respectively.
LIBERO is a language-conditioned multi-task benchmark. For evaluation, we select four suites: LIBERO-Goal (10 tasks), LIBERO-Object (10 tasks), LIBERO-Spatial (10 tasks), and LIBERO-Long (10 tasks), each focusing on the controlled transfer of knowledge related to task goals, objects, spatial information, and long-horizon tasks, respectively. During evaluation, the number of validation trajectories is set to 20.

\begin{figure*}[ht]
\begin{center}
\centerline{\includegraphics[width=\textwidth]{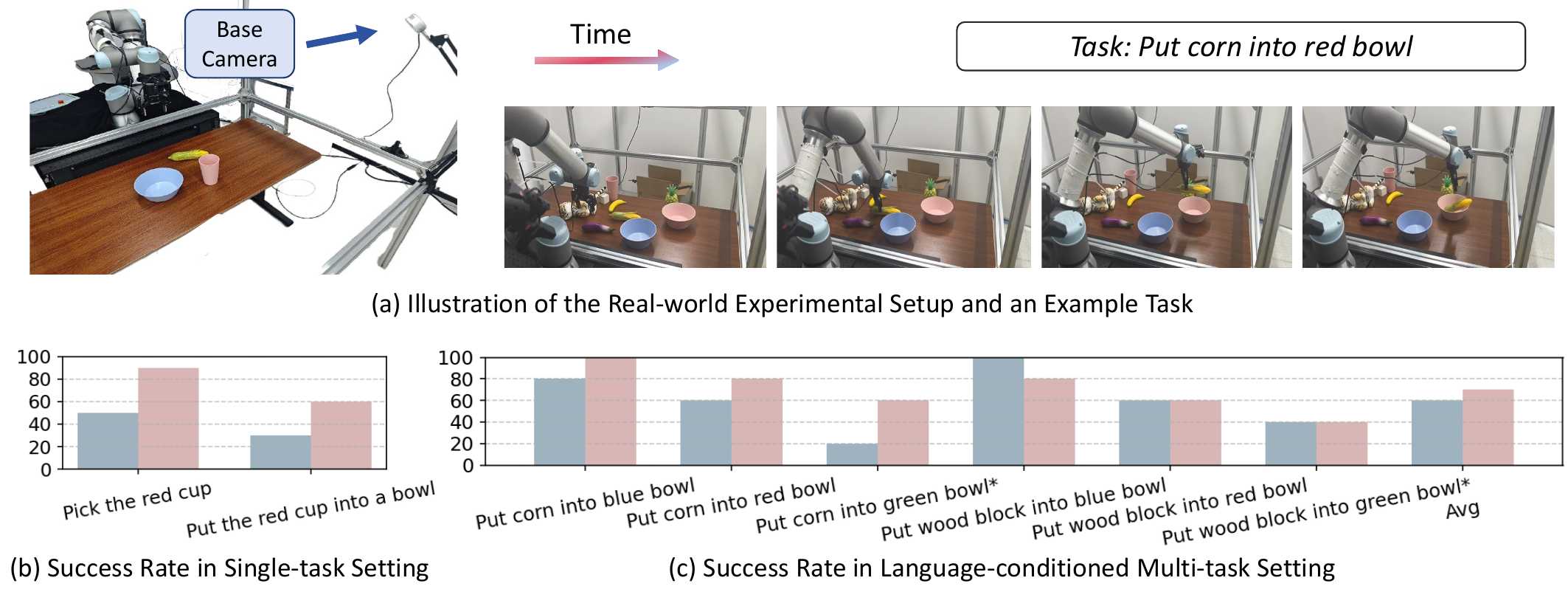}}
\caption{Real-world robot experiments conducted on a tabletop setup with two settings.
(a) Left: the experimental setup.
(a) Right: an example of predicted trajectories alongside policy execution.
(b) and (c): quantitative evaluation results across two settings, where blue denotes the vanilla BC method and red denotes the method with IB.
$^*$ denotes the unseen tasks.
}
\label{fig:real_world}
\end{center}
\vskip -0.25in
\end{figure*}

\textbf{Real-world Evaluation.}
As shown in~\cref{fig:real_world}, our real-world experiments use a 6-DOF UR5 arm equipped with a Robotiq 2F-85 gripper and a RealSense L515 base camera for RGB image capture. 
Following the simulation setup, we evaluate both a single-task setting and a more challenging language-conditioned multi-task setting. The latter introduces increased distractor objects, randomized object positions, and unseen instances during evaluation to assess generalization. 
We design two tabletop manipulation tasks: Pick, where the robot lifts an object from the table, and Put (Pick and Place), where the robot picks up an object and places it into a bowl. 
Demonstrations are collected using a 3D mouse with only the base camera. In the single-task setting, we use 25 demonstrations for Pick and 50 for Pick-and-Place. In the multi-task setting, we collect 800 demonstrations in total, with 200 per task. During evaluation, each task is tested over 10 trajectories.

\textbf{Baselines.} 
In CortexBench, we evaluate four visual imitation learning models: R3M~\cite{nair2023r3m}, Voltron~\cite{karamcheti2023language}, VC-1~\cite{majumdar2023we}, and MPI~\cite{zeng2024learning}. 
Following the original papers, we use pre-trained models with frozen image encoders for downstream tasks. 
Additionally, we introduce two full fine-tuning baselines by replacing the encoders with partially uninitialized ResNet-18~\cite{he2016deep} and ViT-S~\cite{dosovitskiy2020image}, denoted as ResNet and ViT, respectively. 
All methods use the two fusion techniques from~\cref{subsec:ma}: an MLP for spatial fusion and a Temporal Transformer for temporal fusion.
In LIBERO, we implement four vision-language policy networks. One of them uses a spatial fusion approach, which employs ResNet as the image encoder and an MLP as the fusion module, referred to as BC-MLP. The other three use temporal fusion. Following the original paper, we rename them based on the combination of the image encoder and fusion module: BC-RNN, BC-Transformer, and BC-VILT~\cite{liu2024libero}.
The policy head for all methods is fixed as an MLP. 
In real-world evaluation, we adopt VC-1~\cite{majumdar2023we} for the single-task setting and CogAct~\cite{li2024cogact} for the language-conditioned multi-task setting.
Notably, all baselines with IB are referred to as BC+IB.

\textbf{Implementation.} 
In CortexBench, for four partial fine-tuning methods, we train for 100 epochs on each task using the Adam optimizer with a learning rate of 1e-3, a batch size of 512, and weight decay of 1e-4, with learning rate decay applied using a cosine annealing schedule. For two full fine-tuning methods, we train for 50 epochs with a learning rate of 1e-4 and a batch size of 256.
In LIBERO, we train for 50 epochs using the AdamW optimizer with a learning rate of 1e-4 and a batch size of 64, decayed using a cosine annealing schedule.
In real-world evaluation, we train VC-1 for 200 epochs using the Adam optimizer with a learning rate of 1e-3 and a batch size of 512.
CogAct is trained for 8k steps with the AdamW optimizer, using a learning rate of 2e-5 and a batch size of 128.
For BC+IB methods, the model used in MINE consists of a two-layer MLP, with a learning rate of 1e-5. The Lagrange multiplier in~\cref{func:bcib} ranges from 1e-4 to 1e-2 in this work.

\textbf{Model Selection.}
For the single-task benchmark CortexBench, we test the model every 5 or 10 epochs and select the model with the highest success rate.
For the multi-task benchmark LIBERO, we select the model from the final epoch.
For real-world evaluation, we follow the corresponding strategy for each setting as described above.

The appendix provides detailed descriptions of each benchmark (\cref{sec:appendix_benchmarks}), all baselines (\cref{sec:appendix_baselines}), implementation details (\cref{sec:appendix_implementations}), and the rationale behind the model selection (\cref{sec:appendix_model_selection}).

\begin{figure}[ht]
\begin{center}
\centerline{\includegraphics[width=\columnwidth]{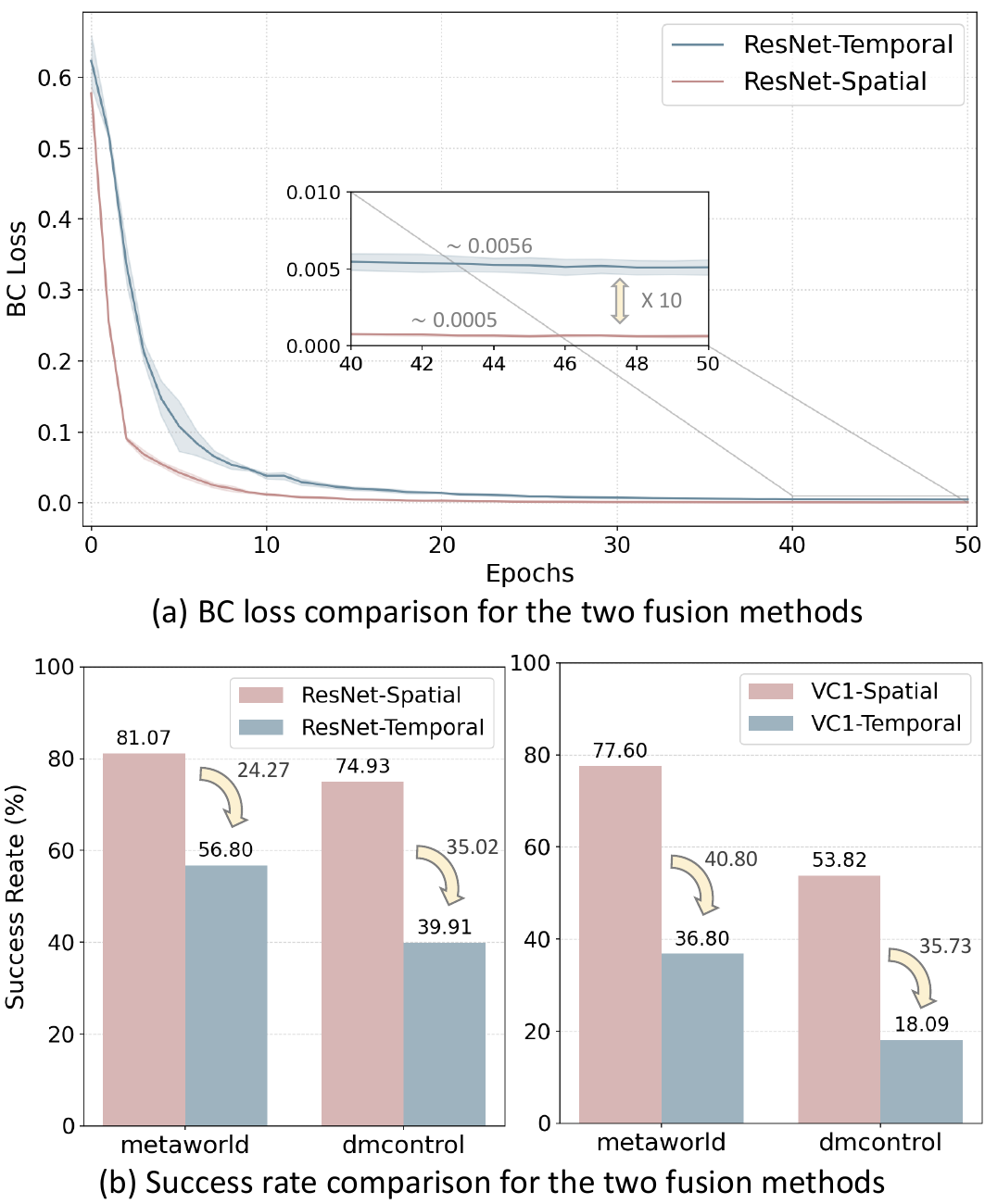}}
\caption{
(a) BC loss variation for ResNet in spatial and temporal fusion methods on the bin-picking task of the Meta-World. (b) Averaged success rates of ResNet and VC1 in spatial and temporal fusion methods across the Meta-World and DMControl.
}
\label{fig:fusion}
\end{center}
\vskip -0.25in
\end{figure}

\subsection{Performance on Cortexbench}

\textbf{The Selection of Fusion Method.} We first evaluate the effectiveness of the two fusion methods in the baselines on CortexBench, with the results shown in~\cref{fig:fusion}.

\textbf{\textit{Finding 1:}}
For simple single-task scenarios, spatial fusion is more efficient and effective than temporal fusion. As shown in~\cref{fig:fusion} (b), the performance of methods with temporal fusion drops significantly. From~\cref{fig:fusion} (a), this can be attributed to the slower loss reduction in methods using temporal fusion, which results in higher loss at the same training epoch. 
Therefore, we focus exclusively on presenting the results for methods employing spatial fusion.

\begin{table*}[ht]
\small
\centering
\caption{Performance in spatial fusion on single-task benchmark CortexBench.
We evaluated 14 tasks across 4 benchmarks using 3 random seeds and reported the average success rate along with the standard deviation. 
$^*$ denotes the use of only a small portion of the original model for feature extraction.
The best performance is highlighted in bold.
}
\vskip -0.05in
\begin{tabular}{
>{\raggedright\arraybackslash}m{4.2cm}
>{\centering\arraybackslash}m{2cm}
>{\centering\arraybackslash}m{1.5cm}
>{\centering\arraybackslash}m{2cm}
>{\centering\arraybackslash}m{1.5cm}
>{\centering\arraybackslash}m{1.5cm}
>{\centering\arraybackslash}m{1.5cm}
}
\toprule
Method              & Image Encoder               & Adroit & Meta-World & DMControl & TriFinger & Avg   \\
\midrule
\multicolumn{7}{l}{\textit{Full Fine-tuning}} \\
\midrule
ResNet~\cite{he2016deep}              & \multirow{2}{*}{ResNet$^*$} & 66.00\scriptsize{$\pm$5.29}  & 81.07\scriptsize{$\pm$1.22}      & 74.93\scriptsize{$\pm$6.21}     & 71.59\scriptsize{$\pm$0.88}     & 73.40 \\
ResNet+IB           &                             & \textbf{72.00}\scriptsize{$\pm$2.00}  & \textbf{83.20}\scriptsize{$\pm$0.80}      & \textbf{84.94}\scriptsize{$\pm$3.54}     & \textbf{72.30}\scriptsize{$\pm$1.76}     & \textbf{78.11} \\
\midrule
ViT~\cite{dosovitskiy2020image}                 & \multirow{2}{*}{ViT$^*$}     & 35.33\scriptsize{$\pm$3.06}  & 31.73\scriptsize{$\pm$1.67}      & 10.41\scriptsize{$\pm$1.21}     & 55.57\scriptsize{$\pm$2.65}     & 33.26 \\
ViT+IB              &                             & \textbf{37.33}\scriptsize{$\pm$4.16}  & \textbf{36.00}\scriptsize{$\pm$6.97}      & \textbf{12.53}\scriptsize{$\pm$2.17}     & \textbf{55.93}\scriptsize{$\pm$2.16}     & \textbf{35.45} \\
\midrule
\multicolumn{7}{l}{\textit{Partial Fine-tuning}} \\
\midrule
R3M~\cite{nair2023r3m}                 & \multirow{2}{*}{ViT-S}      & 25.33\scriptsize{$\pm$6.43}  & 53.07\scriptsize{$\pm$1.67}      & 40.31\scriptsize{$\pm$0.65}     & 59.87\scriptsize{$\pm$0.78}     & 44.65 \\
R3M+IB              &                             & \textbf{27.33}\scriptsize{$\pm$3.06}  & \textbf{54.13}\scriptsize{$\pm$2.44}      & \textbf{41.74}\scriptsize{$\pm$5.54}     & \textbf{60.63}\scriptsize{$\pm$0.53}     & \textbf{45.96} \\
\midrule
Voltron~\cite{karamcheti2023language}              & \multirow{2}{*}{ViT-S}      & 18.67\scriptsize{$\pm$6.11}  & 72.53\scriptsize{$\pm$1.22}      & 25.35\scriptsize{$\pm$2.81}     & 74.21\scriptsize{$\pm$2.61}     & 47.69 \\
Voltron+IB          &                             & \textbf{21.33}\scriptsize{$\pm$5.77}  & \textbf{74.40}\scriptsize{$\pm$3.49}      & \textbf{33.16}\scriptsize{$\pm$6.70}     & \textbf{75.12}\scriptsize{$\pm$2.47}     & \textbf{51.00} \\
\midrule
VC-1~\cite{majumdar2023we}                 & \multirow{2}{*}{ViT-B}      & 24.67\scriptsize{$\pm$7.02}  & 77.60\scriptsize{$\pm$2.88}      & 53.82\scriptsize{$\pm$5.03}     & 72.05\scriptsize{$\pm$2.17}     & 57.04 \\
VC-1+IB             &                             & \textbf{26.00}\scriptsize{$\pm$9.17}  & \textbf{82.40}\scriptsize{$\pm$2.88}      & \textbf{54.93}\scriptsize{$\pm$1.11}     & \textbf{73.80}\scriptsize{$\pm$1.27}     & \textbf{59.28} \\
\midrule
MPI~\cite{zeng2024learning}                  & \multirow{2}{*}{ViT-S}      & 34.67\scriptsize{$\pm$4.16}  & 66.40\scriptsize{$\pm$2.12}      & 59.45\scriptsize{$\pm$1.91}     & 61.91\scriptsize{$\pm$0.57}     & 55.61 \\
MPI+IB              &                             & \textbf{36.67}\scriptsize{$\pm$6.11}  & \textbf{69.33}\scriptsize{$\pm$1.67}      & \textbf{61.41}\scriptsize{$\pm$3.15}     & \textbf{63.34}\scriptsize{$\pm$1.52}     & \textbf{57.69} \\
\bottomrule
\end{tabular}
\label{tab:cortex}
\end{table*}
\begin{table*}[ht]
\small
\centering
\caption{Performance on language-condition multi-task benchmark LIBERO. We evaluated 40 tasks of 4 suites using 3 random seeds and reported the average success rate along with the standard deviation.
S-Trans. denotes Spatial Transformer and T-Trans. denotes Temporal Transformer.
The best performance is bolded.
}
\vskip -0.05in
\begin{tabular}{
>{\raggedright\arraybackslash}m{2cm}
>{\centering\arraybackslash}m{1.5cm}
>{\centering\arraybackslash}m{1.5cm}
>{\centering\arraybackslash}m{1.5cm}
>{\centering\arraybackslash}m{1.5cm}
>{\centering\arraybackslash}m{1.5cm}
>{\centering\arraybackslash}m{1.5cm}
>{\centering\arraybackslash}m{1.5cm}
}
\toprule
Method & Image Encoder & Fuse Module & LIBERO-Goal & LIBERO-Object & LIBERO-Spatial & LIBERO-Long & Avg   \\
\midrule
BC-MLP           & \multirow{2}{*}{ResNet}      & \multirow{2}{*}{MLP}         & 16.50\scriptsize{$\pm$3.97}       & 19.00\scriptsize{$\pm$12.22}         & 29.33\scriptsize{$\pm$9.61}          & 2.33\scriptsize{$\pm$0.76}      & 16.79 \\
BC-MLP+IB        &                              &                              & \textbf{27.67}\scriptsize{$\pm$12.00}       & \textbf{31.50}\scriptsize{$\pm$10.83}         & \textbf{41.00}\scriptsize{$\pm$8.32}          & \textbf{2.67}\scriptsize{$\pm$0.76}      & \textbf{25.71} \\
\midrule
BC-RNN           & \multirow{2}{*}{ResNet}      & \multirow{2}{*}{RNN}         & 15.17\scriptsize{$\pm$10.91}       & 13.33\scriptsize{$\pm$7.91}         & 30.67\scriptsize{$\pm$13.34}          & 2.33\scriptsize{$\pm$0.67}      & 15.38 \\
BC-RNN+IB        &                              &                              & \textbf{26.00}\scriptsize{$\pm$3.50}       & \textbf{17.67}\scriptsize{$\pm$5.77}         & \textbf{35.17}\scriptsize{$\pm$9.45}          & \textbf{3.00}\scriptsize{$\pm$0.17}      & \textbf{20.46} \\
\midrule

BC-Trans.    & \multirow{2}{*}{ResNet}      & \multirow{2}{*}{T-Trans.} & 67.83\scriptsize{$\pm$10.42}       & 41.83\scriptsize{$\pm$1.89}         & 68.00\scriptsize{$\pm$1.00}          & 15.83\scriptsize{$\pm$2.52}     & 48.37 \\
BC-Trans.+IB &                              &                              & \textbf{74.17}\scriptsize{$\pm$5.75}       & \textbf{45.67}\scriptsize{$\pm$4.31}         & \textbf{72.50}\scriptsize{$\pm$10.26}          & \textbf{18.00}\scriptsize{$\pm$6.38}     & \textbf{52.59} \\
\midrule
BC-VILT          & \multirow{2}{*}{S-Trans.} & \multirow{2}{*}{T-Trans.} & 76.17\scriptsize{$\pm$3.01}       & 43.00\scriptsize{$\pm$3.91}         & 67.17\scriptsize{$\pm$2.25}          & 6.50\scriptsize{$\pm$0.87}      & 48.21 \\
BC+VILT+IB       &                              &                              & \textbf{83.83}\scriptsize{$\pm$3.40}       & \textbf{52.00}\scriptsize{$\pm$3.04}         & \textbf{70.67}\scriptsize{$\pm$2.52}          & \textbf{8.67}\scriptsize{$\pm$1.53}      & \textbf{53.79} \\
\bottomrule
\end{tabular}
\label{tab:libero}
\end{table*}

\textbf{Results.} We next report the performance, \textit{i.e.,} success rate, of the baselines and baselines with IB on the single-task benchmark CortexBench in~\cref{tab:cortex} with a full-shot setting. Based on results, we derive the following findings.

\textbf{\textit{Finding 2:}} 
Whether using full fine-tuning or partial fine-tuning, all vanilla BC methods with different visual backbones incorporating IB outperform their vanilla counterparts across the board. 
In some benchmarks, the improvements are substantial. For example, ResNet with IB achieves a 10.01\% improvement on DMControl, and VC-1 with IB shows a 4.80\% improvement on Meta-World. In~\cref{sec:appendix_task-wise_exp}, we report the success rate for each task, where significant improvements can be observed in certain tasks.

\textbf{\textit{Finding 3:}} 
Finding 2 implicitly suggests that the latent representation $Z$ derived from input $X$ is redundant. Therefore, compressing information from input is essential, which can further enhance performance.

\textbf{\textit{Finding 4:}} 
In some benchmarks, particularly Trifinger, the improvement is minimal. We attribute this to the benchmark itself containing very limited redundancy in the visual input, which consists primarily of the robot arm and a single object.

\textbf{\textit{Finding 5:}} 
For simple single-task downstream tasks, full fine-tuning of a simple, uninitialized model (ResNet) is sufficient and may even outperform a pre-trained larger model. However, the latter is more efficient for faster fine-tuning and deployment, and proves to be more effective for more complex tasks~\cite{burns2023makes}.

\subsection{Performance on LIBERO}

\textbf{Results.} We report the performance on the multi-task benchmark LIBERO with a full-shot setting in~\cref{tab:libero}.

\textbf{\textit{Finding 6:}} 
For more complex language-conditioned multi-task scenarios, all baselines with different backbones incorporating IB consistently show performance improvements across all LIBERO benchmarks. For example, BC-VILT achieves large gains of 7.66\% and 9.00\% on LIBERO-Goal and LIBERO-Object, respectively, while BC-RNN shows a significant improvement of 10.83\% on LIBERO-Goal. 
IB proves to be more effective in more complex environments and settings. We attribute this to the difference in task complexity: in CortexBench, the history length is 3, while in LIBERO, it is 10, with LIBERO being a multi-task benchmark and CortexBench being a single-task benchmark. The increased data complexity (task quantity and input information) suggests a higher level of data redundancy, making IB even more effective.

\textbf{\textit{Finding 7:}} 
We observe that in complex multi-task scenarios with more intricate inputs, such as a greater number of input modalities and extended historical information, using the Temporal Transformer in temporal fusion proves to be more effective than both spatial fusion and RNN-based temporal fusion. The evidence lies in the fact that the average success rates of BC-Transformer and BC-VILT are over 30\% higher than those of BC-MLP and BC-RNN.
This is likely because Temporal Transformers excel in handling long-range interactions and capturing dynamic dependencies across time steps, where RNNs and spatial fusion methods may struggle.
This finding, together with Finding 1, underscores the specific scenarios in which each fusion method is most applicable.

\textbf{\textit{Finding 8:}} 
IB is particularly effective for tasks requiring diverse feature extraction, such as distinguishing distinct task objectives or differentiating between various objects, as in LIBERO-Goal and LIBERO-Object. By filtering out irrelevant information, IB facilitates better generalization and more compact representations. 
However, the impact is less pronounced in spatial tasks such as LIBERO-Spatial, which rely heavily on structural information that can be disrupted by excessive compression.
For long-horizon tasks, the main performance bottleneck lies in the lightweight baseline models, whose limited capacity restricts their effectiveness on more complex tasks such as LIBERO-Long.

\begin{figure}[ht]
\begin{center}
\centerline{\includegraphics[width=\columnwidth]{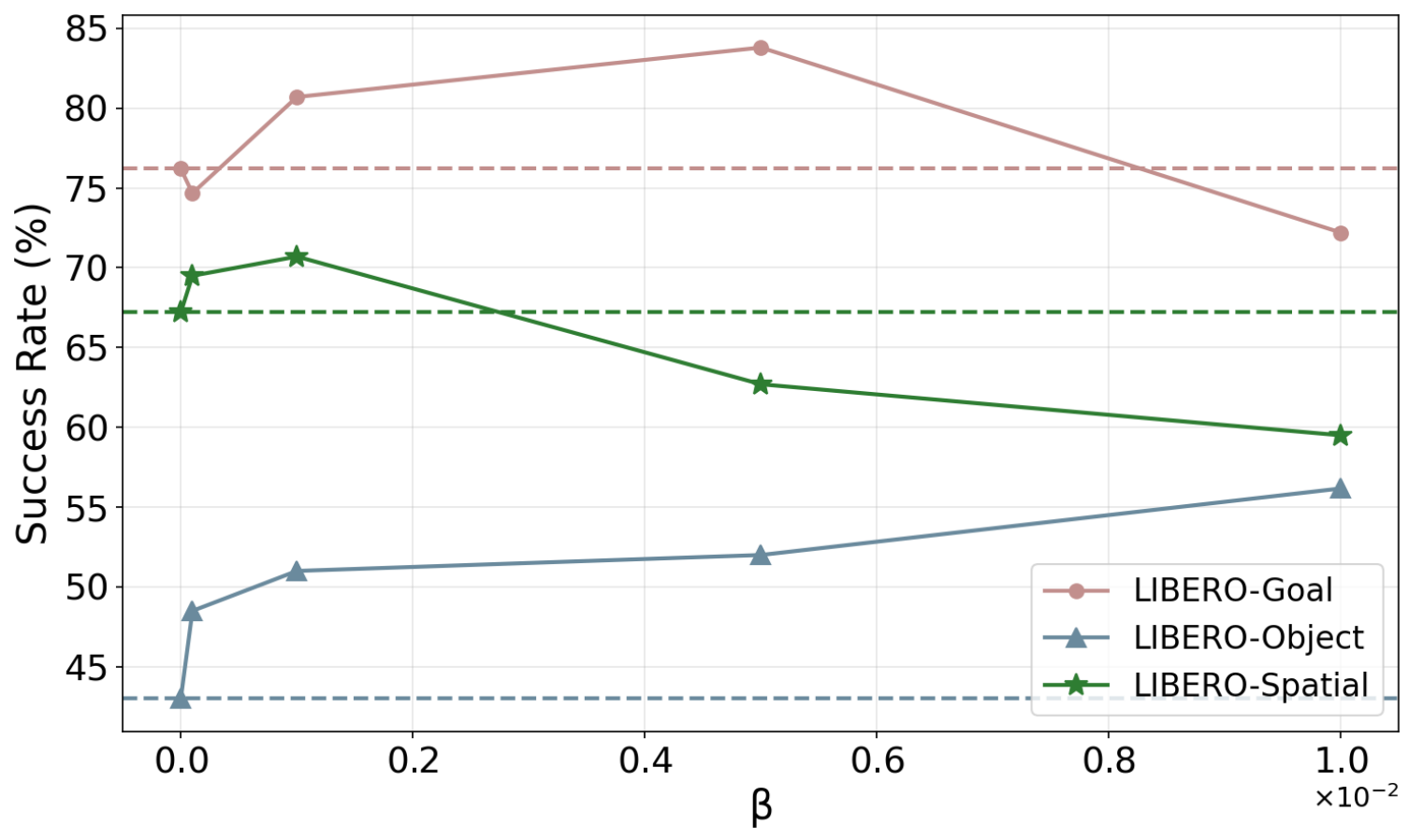}}
\caption{Effect of the Lagrange multiplier $\beta$ in BC-VILT+IB across three suites of LIBERO. When $\beta$=0, the method reduces to vanilla BC-VILT. 
}
\label{fig:beta}
\end{center}
\vskip -0.25in
\end{figure}

\subsection{Performance on Real World Experiments} 

As shown in~\cref{fig:real_world} (b) and (c), incorporating IB consistently improves success rates across both single-task and language-conditioned multi-task real-world settings in most cases. 
In the single-task setting, VC1+IB significantly outperforms VC1 in both the pick and put tasks.
In the more challenging language-conditioned multi-task setting, CogAct+IB consistently outperforms CogAct across most tasks, including unseen object–bowl combinations, demonstrating enhanced generalization capabilities.
These results suggest that reducing redundancy in latent representations leads to more robust grasping and more reliable overall execution in real-world scenarios.

\begin{figure*}[ht]
\begin{center}
\centerline{\includegraphics[width=\textwidth]{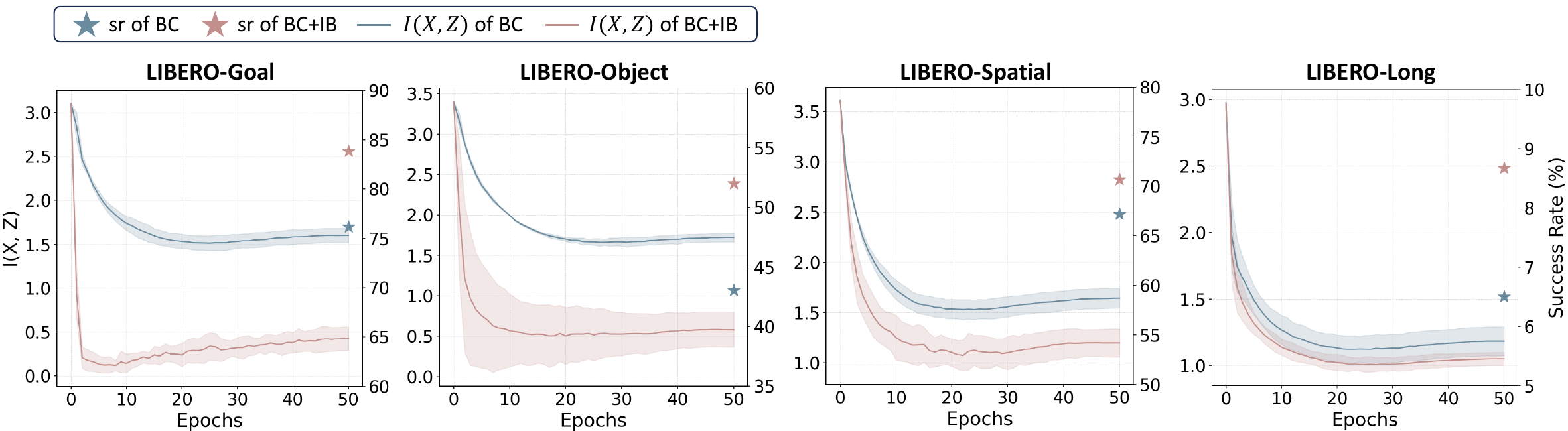}}
\caption{Comparison of vanilla BC and BC+IB on the LIBERO benchmark in terms of success rate (sr) and mutual information. BC-VILT is denoted as BC. BC+IB consistently achieves lower $I(X, Z)$ and higher success rates.
}
\label{fig:mi}
\end{center}
\vskip -0.25in
\end{figure*}

\begin{figure}[ht]
\begin{center}
\centerline{\includegraphics[width=0.99\columnwidth]{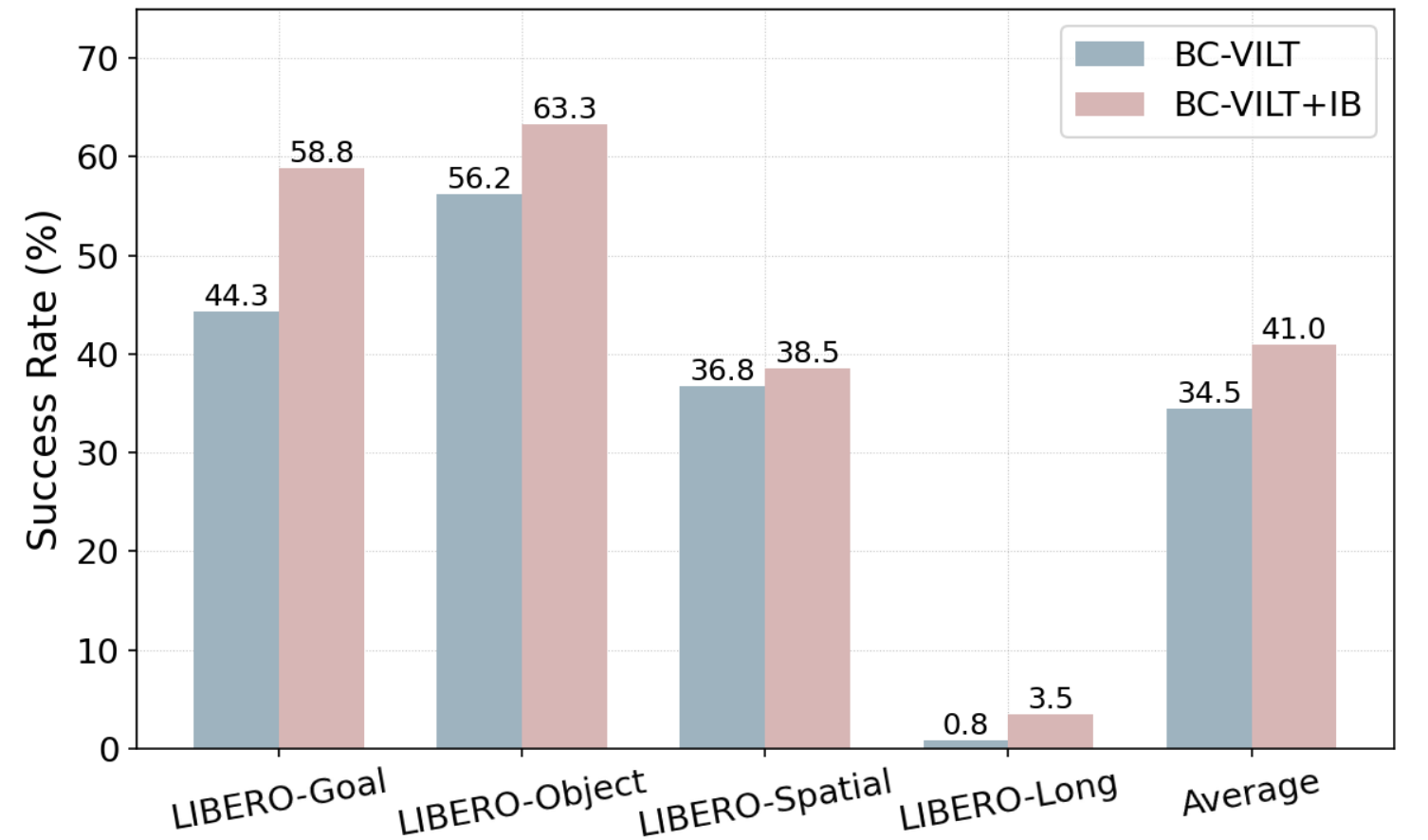}}
\caption{Comparison of the success rates of BC-VILT+IB trained with 10 demonstrations against the vanilla BC-VILT across four LIBERO suites.
}
\label{fig:few-shot}
\end{center}
\vskip -0.25in
\end{figure}

\subsection{More Analysis}

\textbf{Effect of the Lagrange multiplier $\beta$ of~\cref{func:bcib}.} 
This experiment evaluates how incorporating IB enhances performance. Since the MINE model’s parameters are fixed, the key difference between BC+IB and BC lies in the parameter $\beta$, which balances compression and predictive power. For LIBERO experiments, $\beta$ is explored within {1e-4, 1e-3, 5e-3, 1e-2}.
As shown in~\cref{fig:beta}, IB improves performance within a specific $\beta$ range, with a peak observed at an undetermined value. However, across all experiments, $\beta$ around 1e-4 consistently yields stable improvements. 

\textbf{Effect of the Number of Demonstrations.}
We evaluate IB's effectiveness in few-shot settings, as few-shot learning is crucial for fine-tuning on domain-specific tasks in real-world applications.
As shown in~\cref{fig:few-shot}, IB consistently improves performance even with limited data across multiple suites in LIBERO, highlighting its effectiveness in real-world scenarios where data is scarce. This further underscores the potential of IB in improving model generalization in practical settings.

\textbf{Visualizations of $I(X, Z)$.}
As shown in~\cref{fig:mi}, BC+IB achieves a larger reduction in $I(X, Z)$ compared to vanilla BC, leading to improved performance and validating the effectiveness of IB. For example, in LIBERO-Goal, IB reduces $I(X, Z)$ to one-quarter of its original value and yields a 7.7\% increase in success rate.

\section{Limitations and Discussion}
\vspace{-2pt}

While our work provides extensive experimental validation of the effectiveness of IB and the necessity of input redundancy reduction in robotics representation learning, several limitations remain.
First, we do not comprehensively assess the scalability of the IB approach in large-scale robotic models. Most of our experiments are conducted using relatively lightweight architectures, and only the real-world experiments employ CogAct. We have not systematically evaluated more complex models such as vision-language-action (VLA) architectures~\cite{brohan2023rt, song2025accelerating, zhao2025vlas} due to the high computational and time requirements.
Second, we do not explore alternative policy head designs, including transformer-based heads or VLA models~\cite{kim2024openvla, ding2024quar} that remove the explicit policy module and instead treat actions as text tokens. These architectures are left for future investigation.
Third, although our method is evaluated on various benchmarks, its robustness to domain shifts, including changes in environment and task configurations, remains insufficiently studied.
We hope this work inspires future research and advances the development of these methods.

\vspace{-4pt}
\section{Conclusion}
\vspace{-2pt}

In this study, we investigated the redundancy in latent representations for behavior cloning in robot manipulation and introduced the Information Bottleneck (IB) principle to mitigate this issue. 
By incorporating IB, we aimed to filter out redundant information in latent representations while preserving task-relevant features. 
Extensive experiments across various visual representation learning methods on CortexBench and LIBERO benchmark revealed insightful findings and demonstrated that IB consistently improves performance across diverse tasks and architectures. 
We hope this work will inspire future research to further integrate information-theoretic principles into robotics, not only as an optimization tool, but also as a framework for theoretical understanding and model design.

\section*{Acknowledgements}
This work was supported by the National Natural Science Foundation of China (Grant No. U21A20485), the National Science and Technology Innovation 2030 – Major Project (Grant No. 2022ZD0208800), the NSFC General Program (Grant No. 62176215, 62436005), and the Fundamental Research Funds for Xi’an Jiaotong University (Grant No. xzy022024012).

\section*{Impact Statement}
This paper presents work whose goal is to advance the field of Machine Learning in Robotics. There are many potential societal consequences of our work, none of which we feel must be specifically highlighted here.


\bibliography{main}

\begin{thebibliography}{71}
\providecommand{\natexlab}[1]{#1}
\providecommand{\url}[1]{\texttt{#1}}
\expandafter\ifx\csname urlstyle\endcsname\relax
  \providecommand{\doi}[1]{doi: #1}\else
  \providecommand{\doi}{doi: \begingroup \urlstyle{rm}\Url}\fi

\bibitem[Amjad \& Geiger(2019)Amjad and Geiger]{amjad2019learning}
Amjad, R.~A. and Geiger, B.~C.
\newblock Learning representations for neural network-based classification using the information bottleneck principle.
\newblock \emph{IEEE transactions on pattern analysis and machine intelligence}, 42\penalty0 (9):\penalty0 2225--2239, 2019.

\bibitem[Argall et~al.(2009)Argall, Chernova, Veloso, and Browning]{argall2009survey}
Argall, B.~D., Chernova, S., Veloso, M., and Browning, B.
\newblock A survey of robot learning from demonstration.
\newblock \emph{Robotics and autonomous systems}, 57\penalty0 (5):\penalty0 469--483, 2009.

\bibitem[Bai et~al.(2021)Bai, Wang, Han, Garg, Hao, Liu, and Wang]{bai2021dynamic}
Bai, C., Wang, L., Han, L., Garg, A., Hao, J., Liu, P., and Wang, Z.
\newblock Dynamic bottleneck for robust self-supervised exploration.
\newblock \emph{Advances in Neural Information Processing Systems}, 34:\penalty0 17007--17020, 2021.

\bibitem[Bain \& Sammut(1995)Bain and Sammut]{bain1995framework}
Bain, M. and Sammut, C.
\newblock A framework for behavioural cloning.
\newblock In \emph{Machine Intelligence 15}, pp.\  103--129, 1995.

\bibitem[Bardera et~al.(2009)Bardera, Rigau, Boada, Feixas, and Sbert]{bardera2009image}
Bardera, A., Rigau, J., Boada, I., Feixas, M., and Sbert, M.
\newblock Image segmentation using information bottleneck method.
\newblock \emph{IEEE Transactions on Image Processing}, 18\penalty0 (7):\penalty0 1601--1612, 2009.

\bibitem[Belghazi et~al.(2018)Belghazi, Baratin, Rajeshwar, Ozair, Bengio, Courville, and Hjelm]{belghazi2018mutual}
Belghazi, M.~I., Baratin, A., Rajeshwar, S., Ozair, S., Bengio, Y., Courville, A., and Hjelm, D.
\newblock Mutual information neural estimation.
\newblock In \emph{International conference on machine learning}, pp.\  531--540. PMLR, 2018.

\bibitem[Brohan et~al.(2023)Brohan, Brown, Carbajal, Chebotar, Chen, Choromanski, Ding, Driess, Dubey, Finn, et~al.]{brohan2023rt}
Brohan, A., Brown, N., Carbajal, J., Chebotar, Y., Chen, X., Choromanski, K., Ding, T., Driess, D., Dubey, A., Finn, C., et~al.
\newblock Rt-2: Vision-language-action models transfer web knowledge to robotic control.
\newblock In \emph{Conference on Robot Learning}, 2023.

\bibitem[Burns et~al.(2023)Burns, Witzel, Hamid, Yu, Finn, and Hausman]{burns2023makes}
Burns, K., Witzel, Z., Hamid, J.~I., Yu, T., Finn, C., and Hausman, K.
\newblock What makes pre-trained visual representations successful for robust manipulation?
\newblock \emph{arXiv preprint arXiv:2312.12444}, 2023.

\bibitem[Cheang et~al.(2024)Cheang, Chen, Jing, Kong, Li, Li, Liu, Wu, Xu, Yang, et~al.]{cheang2024gr}
Cheang, C.-L., Chen, G., Jing, Y., Kong, T., Li, H., Li, Y., Liu, Y., Wu, H., Xu, J., Yang, Y., et~al.
\newblock Gr-2: A generative video-language-action model with web-scale knowledge for robot manipulation.
\newblock \emph{arXiv preprint arXiv:2410.06158}, 2024.

\bibitem[Chi et~al.(2023)Chi, Xu, Feng, Cousineau, Du, Burchfiel, Tedrake, and Song]{chi2023diffusion}
Chi, C., Xu, Z., Feng, S., Cousineau, E., Du, Y., Burchfiel, B., Tedrake, R., and Song, S.
\newblock Diffusion policy: Visuomotor policy learning via action diffusion.
\newblock \emph{The International Journal of Robotics Research}, pp.\  02783649241273668, 2023.

\bibitem[Cover(1999)]{AEP}
Cover, T.~M.
\newblock \emph{Elements of information theory}.
\newblock John Wiley \& Sons, 1999.

\bibitem[Cui et~al.(2025)Cui, Ding, Song, Bai, Tong, Ge, Suo, Zhou, Liu, Jia, et~al.]{cui2025openhelix}
Cui, C., Ding, P., Song, W., Bai, S., Tong, X., Ge, Z., Suo, R., Zhou, W., Liu, Y., Jia, B., et~al.
\newblock Openhelix: A short survey, empirical analysis, and open-source dual-system vla model for robotic manipulation.
\newblock \emph{arXiv preprint arXiv:2505.03912}, 2025.

\bibitem[Deng et~al.(2009)Deng, Dong, Socher, Li, Li, and Fei-Fei]{deng2009imagenet}
Deng, J., Dong, W., Socher, R., Li, L.-J., Li, K., and Fei-Fei, L.
\newblock Imagenet: A large-scale hierarchical image database.
\newblock In \emph{2009 IEEE conference on computer vision and pattern recognition}, pp.\  248--255. Ieee, 2009.

\bibitem[Ding et~al.(2024)Ding, Zhao, Zhang, Song, Zhang, Huang, Yang, and Wang]{ding2024quar}
Ding, P., Zhao, H., Zhang, W., Song, W., Zhang, M., Huang, S., Yang, N., and Wang, D.
\newblock Quar-vla: Vision-language-action model for quadruped robots.
\newblock In \emph{European Conference on Computer Vision}, pp.\  352--367. Springer, 2024.

\bibitem[Donsker \& Varadhan(1983)Donsker and Varadhan]{donsker1983asymptotic}
Donsker, M.~D. and Varadhan, S.~S.
\newblock Asymptotic evaluation of certain markov process expectations for large time. iv.
\newblock \emph{Communications on pure and applied mathematics}, 36\penalty0 (2):\penalty0 183--212, 1983.

\bibitem[Dosovitskiy(2021)]{dosovitskiy2020image}
Dosovitskiy, A.
\newblock An image is worth 16x16 words: Transformers for image recognition at scale.
\newblock In \emph{International Conference on Learning Representations}, 2021.

\bibitem[Federici et~al.(2019)Federici, Dutta, Forr{\'e}, Kushman, and Akata]{federici2019learning}
Federici, M., Dutta, A., Forr{\'e}, P., Kushman, N., and Akata, Z.
\newblock Learning robust representations via multi-view information bottleneck.
\newblock In \emph{International Conference on Learning Representations}, 2019.

\bibitem[Gong et~al.(2024)Gong, Ding, Lyu, Huang, Sun, Zhao, Fan, and Wang]{gong2024carp}
Gong, Z., Ding, P., Lyu, S., Huang, S., Sun, M., Zhao, W., Fan, Z., and Wang, D.
\newblock Carp: Visuomotor policy learning via coarse-to-fine autoregressive prediction.
\newblock \emph{arXiv preprint arXiv:2412.06782}, 2024.

\bibitem[Goyal et~al.(2023)Goyal, Xu, Guo, Blukis, Chao, and Fox]{goyal2023rvt}
Goyal, A., Xu, J., Guo, Y., Blukis, V., Chao, Y.-W., and Fox, D.
\newblock Rvt: Robotic view transformer for 3d object manipulation.
\newblock In \emph{Conference on Robot Learning}, pp.\  694--710. PMLR, 2023.

\bibitem[Grauman et~al.(2022)Grauman, Westbury, Byrne, Chavis, Furnari, Girdhar, Hamburger, Jiang, Liu, Liu, et~al.]{grauman2022ego4d}
Grauman, K., Westbury, A., Byrne, E., Chavis, Z., Furnari, A., Girdhar, R., Hamburger, J., Jiang, H., Liu, M., Liu, X., et~al.
\newblock Ego4d: Around the world in 3,000 hours of egocentric video.
\newblock In \emph{Proceedings of the IEEE/CVF Conference on Computer Vision and Pattern Recognition}, pp.\  18995--19012, 2022.

\bibitem[He et~al.(2024)He, Wu, Bai, Lai, Wang, Pan, Hu, and Zhang]{he2024bridging}
He, H., Wu, P., Bai, C., Lai, H., Wang, L., Pan, L., Hu, X., and Zhang, W.
\newblock Bridging the sim-to-real gap from the information bottleneck perspective.
\newblock In \emph{8th Annual Conference on Robot Learning}, 2024.

\bibitem[He et~al.(2016)He, Zhang, Ren, and Sun]{he2016deep}
He, K., Zhang, X., Ren, S., and Sun, J.
\newblock Deep residual learning for image recognition.
\newblock In \emph{Proceedings of the IEEE conference on computer vision and pattern recognition}, pp.\  770--778, 2016.

\bibitem[He et~al.(2022)He, Chen, Xie, Li, Doll{\'a}r, and Girshick]{he2022masked}
He, K., Chen, X., Xie, S., Li, Y., Doll{\'a}r, P., and Girshick, R.
\newblock Masked autoencoders are scalable vision learners.
\newblock In \emph{Proceedings of the IEEE/CVF conference on computer vision and pattern recognition}, pp.\  16000--16009, 2022.

\bibitem[Hu et~al.(2024)Hu, Guo, Wang, Chen, Wang, Zhang, Sreenath, Lu, and Chen]{hu2024video}
Hu, Y., Guo, Y., Wang, P., Chen, X., Wang, Y.-J., Zhang, J., Sreenath, K., Lu, C., and Chen, J.
\newblock Video prediction policy: A generalist robot policy with predictive visual representations.
\newblock \emph{arXiv preprint arXiv:2412.14803}, 2024.

\bibitem[Hu et~al.(2025)Hu, Guo, Wang, Chen, Wang, Zhang, Sreenath, Lu, and Chen]{hu2025video}
Hu, Y., Guo, Y., Wang, P., Chen, X., Wang, Y.-J., Zhang, J., Sreenath, K., Lu, C., and Chen, J.
\newblock Video prediction policy: A generalist robot policy with predictive visual representations.
\newblock In \emph{International conference on machine learning}, 2025.

\bibitem[Jang et~al.(2022)Jang, Irpan, Khansari, Kappler, Ebert, Lynch, Levine, and Finn]{jang2022bc}
Jang, E., Irpan, A., Khansari, M., Kappler, D., Ebert, F., Lynch, C., Levine, S., and Finn, C.
\newblock Bc-z: Zero-shot task generalization with robotic imitation learning.
\newblock In \emph{Conference on Robot Learning}, pp.\  991--1002. PMLR, 2022.

\bibitem[Jeon et~al.(2021)Jeon, Lee, Pyeon, and Kim]{jeon2021ib}
Jeon, I., Lee, W., Pyeon, M., and Kim, G.
\newblock Ib-gan: Disentangled representation learning with information bottleneck generative adversarial networks.
\newblock In \emph{Proceedings of the AAAI conference on artificial intelligence}, volume~35, pp.\  7926--7934, 2021.

\bibitem[Jia et~al.(2024)Jia, Thumuluri, Liu, Chen, Huang, and Su]{jia2024chain}
Jia, Z., Thumuluri, V., Liu, F., Chen, L., Huang, Z., and Su, H.
\newblock Chain-of-thought predictive control.
\newblock In \emph{Forty-first International Conference on Machine Learning}, 2024.

\bibitem[Karamcheti et~al.(2023)Karamcheti, Nair, Chen, Kollar, Finn, Sadigh, and Liang]{karamcheti2023language}
Karamcheti, S., Nair, S., Chen, A.~S., Kollar, T., Finn, C., Sadigh, D., and Liang, P.
\newblock Language-driven representation learning for robotics.
\newblock In \emph{Robotics: Science and Systems}, 2023.

\bibitem[Kawaguchi et~al.(2023)Kawaguchi, Deng, Ji, and Huang]{IB2}
Kawaguchi, K., Deng, Z., Ji, X., and Huang, J.
\newblock How does information bottleneck help deep learning?
\newblock In \emph{International Conference on Machine Learning}, pp.\  16049--16096. PMLR, 2023.

\bibitem[Kim et~al.(2024)Kim, Pertsch, Karamcheti, Xiao, Balakrishna, Nair, Rafailov, Foster, Lam, Sanketi, et~al.]{kim2024openvla}
Kim, M.~J., Pertsch, K., Karamcheti, S., Xiao, T., Balakrishna, A., Nair, S., Rafailov, R., Foster, E., Lam, G., Sanketi, P., et~al.
\newblock Openvla: An open-source vision-language-action model.
\newblock \emph{arXiv preprint arXiv:2406.09246}, 2024.

\bibitem[Kim et~al.(2019)Kim, Nam, Kim, Kim, and Kim]{kim2019curiosity}
Kim, Y., Nam, W., Kim, H., Kim, J.-H., and Kim, G.
\newblock Curiosity-bottleneck: Exploration by distilling task-specific novelty.
\newblock In \emph{International conference on machine learning}, pp.\  3379--3388. PMLR, 2019.

\bibitem[Lee et~al.(2021)Lee, Choi, Mok, and Yoon]{lee2021reducing}
Lee, J., Choi, J., Mok, J., and Yoon, S.
\newblock Reducing information bottleneck for weakly supervised semantic segmentation.
\newblock \emph{Advances in neural information processing systems}, 34:\penalty0 27408--27421, 2021.

\bibitem[Li et~al.(2024{\natexlab{a}})Li, Liang, Wang, Luo, Chen, Liao, Wei, Deng, Xu, Zhang, et~al.]{li2024cogact}
Li, Q., Liang, Y., Wang, Z., Luo, L., Chen, X., Liao, M., Wei, F., Deng, Y., Xu, S., Zhang, Y., et~al.
\newblock Cogact: A foundational vision-language-action model for synergizing cognition and action in robotic manipulation.
\newblock \emph{arXiv preprint arXiv:2411.19650}, 2024{\natexlab{a}}.

\bibitem[Li et~al.(2024{\natexlab{b}})Li, Liu, Zhang, Yu, Xu, Wu, Cheang, Jing, Zhang, Liu, et~al.]{li2023vision}
Li, X., Liu, M., Zhang, H., Yu, C., Xu, J., Wu, H., Cheang, C., Jing, Y., Zhang, W., Liu, H., et~al.
\newblock Vision-language foundation models as effective robot imitators.
\newblock In \emph{International Conference on Learning Representations}, 2024{\natexlab{b}}.

\bibitem[Liu et~al.(2024)Liu, Zhu, Gao, Feng, Liu, Zhu, and Stone]{liu2024libero}
Liu, B., Zhu, Y., Gao, C., Feng, Y., Liu, Q., Zhu, Y., and Stone, P.
\newblock Libero: Benchmarking knowledge transfer for lifelong robot learning.
\newblock \emph{Advances in Neural Information Processing Systems}, 36, 2024.

\bibitem[Ma et~al.(2024)Ma, Zhou, Wang, Qiu, and Liang]{ma2024contrastive}
Ma, T., Zhou, J., Wang, Z., Qiu, R., and Liang, J.
\newblock Contrastive imitation learning for language-guided multi-task robotic manipulation.
\newblock In \emph{Conference on Robot Learning}, 2024.

\bibitem[Majumdar et~al.(2023)Majumdar, Yadav, Arnaud, Ma, Chen, Silwal, Jain, Berges, Wu, Vakil, et~al.]{majumdar2023we}
Majumdar, A., Yadav, K., Arnaud, S., Ma, J., Chen, C., Silwal, S., Jain, A., Berges, V.-P., Wu, T., Vakil, J., et~al.
\newblock Where are we in the search for an artificial visual cortex for embodied intelligence?
\newblock \emph{Advances in Neural Information Processing Systems}, 36:\penalty0 655--677, 2023.

\bibitem[Nair et~al.(2023)Nair, Rajeswaran, Kumar, Finn, and Gupta]{nair2023r3m}
Nair, S., Rajeswaran, A., Kumar, V., Finn, C., and Gupta, A.
\newblock R3m: A universal visual representation for robot manipulation.
\newblock In \emph{Conference on Robot Learning}, pp.\  892--909. PMLR, 2023.

\bibitem[Pacelli \& Majumdar(2020)Pacelli and Majumdar]{pacelli2020learning}
Pacelli, V. and Majumdar, A.
\newblock Learning task-driven control policies via information bottlenecks.
\newblock In \emph{Robotics: Science and Systems (RSS)}, 2020.

\bibitem[Pearce \& Zhu(2022)Pearce and Zhu]{pearce2022counter}
Pearce, T. and Zhu, J.
\newblock Counter-strike deathmatch with large-scale behavioural cloning.
\newblock In \emph{2022 IEEE Conference on Games (CoG)}, pp.\  104--111. IEEE, 2022.

\bibitem[Perez et~al.(2018)Perez, Strub, De~Vries, Dumoulin, and Courville]{perez2018film}
Perez, E., Strub, F., De~Vries, H., Dumoulin, V., and Courville, A.
\newblock Film: Visual reasoning with a general conditioning layer.
\newblock In \emph{Proceedings of the AAAI conference on artificial intelligence}, volume~32, 2018.

\bibitem[Pomerleau(1991)]{pomerleau1991efficient}
Pomerleau, D.~A.
\newblock Efficient training of artificial neural networks for autonomous navigation.
\newblock \emph{Neural computation}, 3\penalty0 (1):\penalty0 88--97, 1991.

\bibitem[Radosavovic et~al.(2023)Radosavovic, Xiao, James, Abbeel, Malik, and Darrell]{radosavovic2023real}
Radosavovic, I., Xiao, T., James, S., Abbeel, P., Malik, J., and Darrell, T.
\newblock Real-world robot learning with masked visual pre-training.
\newblock In \emph{Conference on Robot Learning}, pp.\  416--426. PMLR, 2023.

\bibitem[Rajeswaran et~al.(2018)Rajeswaran, Kumar, Gupta, Vezzani, Schulman, Todorov, and Levine]{rajeswaran2018learning}
Rajeswaran, A., Kumar, V., Gupta, A., Vezzani, G., Schulman, J., Todorov, E., and Levine, S.
\newblock Learning complex dexterous manipulation with deep reinforcement learning and demonstrations.
\newblock \emph{Robotics: Science and Systems}, 2018.

\bibitem[Reuss et~al.(2024)Reuss, Ya{\u{g}}murlu, Wenzel, and Lioutikov]{reuss2024multimodal}
Reuss, M., Ya{\u{g}}murlu, {\"O}.~E., Wenzel, F., and Lioutikov, R.
\newblock Multimodal diffusion transformer: Learning versatile behavior from multimodal goals.
\newblock In \emph{First Workshop on Vision-Language Models for Navigation and Manipulation at ICRA 2024}, 2024.

\bibitem[Saxena et~al.(2025)Saxena, Bronars, Arachchige, Wang, Shin, Nasiriany, Mandlekar, and Xu]{saxena2025matters}
Saxena, V., Bronars, M., Arachchige, N.~R., Wang, K., Shin, W.~C., Nasiriany, S., Mandlekar, A., and Xu, D.
\newblock What matters in learning from large-scale datasets for robot manipulation.
\newblock In \emph{International Conference on Learning Representations}, 2025.

\bibitem[Shwartz-Ziv et~al.(2019)Shwartz-Ziv, Painsky, and Tishby]{IB1}
Shwartz-Ziv, R., Painsky, A., and Tishby, N.
\newblock {REPRESENTATION} {COMPRESSION} {AND} {GENERALIZATION} {IN} {DEEP} {NEURAL} {NETWORKS}, 2019.
\newblock URL \url{https://openreview.net/forum?id=SkeL6sCqK7}.

\bibitem[Song et~al.(2025)Song, Chen, Ding, Zhao, Zhao, Zhong, Ge, Ma, and Li]{song2025accelerating}
Song, W., Chen, J., Ding, P., Zhao, H., Zhao, W., Zhong, Z., Ge, Z., Ma, J., and Li, H.
\newblock Accelerating vision-language-action model integrated with action chunking via parallel decoding.
\newblock \emph{arXiv preprint arXiv:2503.02310}, 2025.

\bibitem[Tassa et~al.(2018)Tassa, Doron, Muldal, Erez, Li, Casas, Budden, Abdolmaleki, Merel, Lefrancq, et~al.]{tassa2018deepmind}
Tassa, Y., Doron, Y., Muldal, A., Erez, T., Li, Y., Casas, D. d.~L., Budden, D., Abdolmaleki, A., Merel, J., Lefrancq, A., et~al.
\newblock Deepmind control suite.
\newblock \emph{arXiv preprint arXiv:1801.00690}, 2018.

\bibitem[Tian et~al.(2025)Tian, Yang, Zeng, Wang, Lin, Dong, and Pang]{tian2025predictive}
Tian, Y., Yang, S., Zeng, J., Wang, P., Lin, D., Dong, H., and Pang, J.
\newblock Predictive inverse dynamics models are scalable learners for robotic manipulation.
\newblock In \emph{International Conference on Learning Representations}, 2025.

\bibitem[Tishby et~al.(1999)Tishby, Pereira, and Bialek]{tishby2000information}
Tishby, N., Pereira, F.~C., and Bialek, W.
\newblock The information bottleneck method.
\newblock \emph{arXiv preprint physics/0004057}, 1999.

\bibitem[Torabi et~al.(2018)Torabi, Warnell, and Stone]{torabi2018behavioral}
Torabi, F., Warnell, G., and Stone, P.
\newblock Behavioral cloning from observation.
\newblock In \emph{Proceedings of the 27th International Joint Conference on Artificial Intelligence}, pp.\  4950--4957, 2018.

\bibitem[Vaswani(2017)]{vaswani2017attention}
Vaswani, A.
\newblock Attention is all you need.
\newblock \emph{Advances in Neural Information Processing Systems}, 2017.

\bibitem[Wan et~al.(2021)Wan, Zhang, Zhu, and Hu]{wan2021multi}
Wan, Z., Zhang, C., Zhu, P., and Hu, Q.
\newblock Multi-view information-bottleneck representation learning.
\newblock In \emph{Proceedings of the AAAI conference on artificial intelligence}, volume~35, pp.\  10085--10092, 2021.

\bibitem[Wang et~al.(2024)Wang, Chen, Zhao, and He]{wang2024scaling}
Wang, L., Chen, X., Zhao, J., and He, K.
\newblock Scaling proprioceptive-visual learning with heterogeneous pre-trained transformers.
\newblock In \emph{Advances in neural information processing systems}, 2024.

\bibitem[Wen et~al.(2020)Wen, Lin, Darrell, Jayaraman, and Gao]{wen2020fighting}
Wen, C., Lin, J., Darrell, T., Jayaraman, D., and Gao, Y.
\newblock Fighting copycat agents in behavioral cloning from observation histories.
\newblock \emph{Advances in Neural Information Processing Systems}, 33:\penalty0 2564--2575, 2020.

\bibitem[Wen et~al.(2024{\natexlab{a}})Wen, Lin, So, Chen, Dou, Gao, and Abbeel]{wen2023any}
Wen, C., Lin, X., So, J., Chen, K., Dou, Q., Gao, Y., and Abbeel, P.
\newblock Any-point trajectory modeling for policy learning.
\newblock In \emph{Robotics: Science and Systems}, 2024{\natexlab{a}}.

\bibitem[Wen et~al.(2024{\natexlab{b}})Wen, Zhu, Zhu, Tang, Li, Zhou, Li, Liu, Peng, Shen, et~al.]{wen2024diffusion}
Wen, J., Zhu, M., Zhu, Y., Tang, Z., Li, J., Zhou, Z., Li, C., Liu, X., Peng, Y., Shen, C., et~al.
\newblock Diffusion-vla: Scaling robot foundation models via unified diffusion and autoregression.
\newblock \emph{arXiv preprint arXiv:2412.03293}, 2024{\natexlab{b}}.

\bibitem[Wen et~al.(2025)Wen, Zhu, Li, Tang, Shen, and Feng]{wen2025dexvla}
Wen, J., Zhu, Y., Li, J., Tang, Z., Shen, C., and Feng, F.
\newblock Dexvla: Vision-language model with plug-in diffusion expert for general robot control.
\newblock \emph{arXiv preprint arXiv:2502.05855}, 2025.

\bibitem[Wu et~al.(2024)Wu, Jing, Cheang, Chen, Xu, Li, Liu, Li, and Kong]{wu2023unleashing}
Wu, H., Jing, Y., Cheang, C., Chen, G., Xu, J., Li, X., Liu, M., Li, H., and Kong, T.
\newblock Unleashing large-scale video generative pre-training for visual robot manipulation.
\newblock In \emph{International Conference on Learning Representations}, 2024.

\bibitem[Wuthrich et~al.(2021)Wuthrich, Widmaier, Grimminger, Joshi, Agrawal, Hammoud, Khadiv, Bogdanovic, Berenz, Viereck, et~al.]{wuthrich2021trifinger}
Wuthrich, M., Widmaier, F., Grimminger, F., Joshi, S., Agrawal, V., Hammoud, B., Khadiv, M., Bogdanovic, M., Berenz, V., Viereck, J., et~al.
\newblock Trifinger: An open-source robot for learning dexterity.
\newblock In \emph{Conference on Robot Learning}, pp.\  1871--1882. PMLR, 2021.

\bibitem[Yu et~al.(2020)Yu, Quillen, He, Julian, Hausman, Finn, and Levine]{yu2020meta}
Yu, T., Quillen, D., He, Z., Julian, R., Hausman, K., Finn, C., and Levine, S.
\newblock Meta-world: A benchmark and evaluation for multi-task and meta reinforcement learning.
\newblock In \emph{Conference on robot learning}, pp.\  1094--1100. PMLR, 2020.

\bibitem[Zawalski et~al.(2024)Zawalski, Chen, Pertsch, Mees, Finn, and Levine]{zawalski2024robotic}
Zawalski, M., Chen, W., Pertsch, K., Mees, O., Finn, C., and Levine, S.
\newblock Robotic control via embodied chain-of-thought reasoning.
\newblock In \emph{8th Annual Conference on Robot Learning}, 2024.

\bibitem[Ze et~al.(2024)Ze, Zhang, Zhang, Hu, Wang, and Xu]{ze20243d}
Ze, Y., Zhang, G., Zhang, K., Hu, C., Wang, M., and Xu, H.
\newblock 3d diffusion policy: Generalizable visuomotor policy learning via simple 3d representations.
\newblock In \emph{ICRA 2024 Workshop on 3D Visual Representations for Robot Manipulation}, 2024.

\bibitem[Zeng et~al.(2024)Zeng, Bu, Wang, Xia, Chen, Dong, Song, Wang, Hu, Luo, et~al.]{zeng2024learning}
Zeng, J., Bu, Q., Wang, B., Xia, W., Chen, L., Dong, H., Song, H., Wang, D., Hu, D., Luo, P., et~al.
\newblock Learning manipulation by predicting interaction.
\newblock In \emph{Robotics: Science and Systems}, 2024.

\bibitem[Zhang et~al.(2025{\natexlab{a}})Zhang, Ding, Lyu, Peng, and Wang]{zhang2025gevrm}
Zhang, H., Ding, P., Lyu, S., Peng, Y., and Wang, D.
\newblock Gevrm: Goal-expressive video generation model for robust visual manipulation.
\newblock In \emph{International Conference on Learning Representations}, 2025{\natexlab{a}}.

\bibitem[Zhang et~al.(2024)Zhang, Guo, Chen, Wang, Hu, Shi, and Chen]{zhang2024hirt}
Zhang, J., Guo, Y., Chen, X., Wang, Y.-J., Hu, Y., Shi, C., and Chen, J.
\newblock Hirt: Enhancing robotic control with hierarchical robot transformers.
\newblock In \emph{8th Annual Conference on Robot Learning}, 2024.

\bibitem[Zhang et~al.(2025{\natexlab{b}})Zhang, Guo, Hu, Chen, Zhu, and Chen]{zhang2025up}
Zhang, J., Guo, Y., Hu, Y., Chen, X., Zhu, X., and Chen, J.
\newblock Up-vla: A unified understanding and prediction model for embodied agent.
\newblock \emph{arXiv preprint arXiv:2501.18867}, 2025{\natexlab{b}}.

\bibitem[Zhao et~al.(2025)Zhao, Ding, Zhang, Gong, Bai, Zhao, and Wang]{zhao2025vlas}
Zhao, W., Ding, P., Zhang, M., Gong, Z., Bai, S., Zhao, H., and Wang, D.
\newblock Vlas: Vision-language-action model with speech instructions for customized robot manipulation.
\newblock In \emph{International Conference on Learning Representations}, 2025.

\bibitem[Zhu et~al.(2024)Zhu, Yang, Wang, Yang, Wang, and He]{zhu2024spa}
Zhu, H., Yang, H., Wang, Y., Yang, J., Wang, L., and He, T.
\newblock Spa: 3d spatial-awareness enables effective embodied representation.
\newblock \emph{arXiv preprint arXiv:2410.08208}, 2024.

\end{thebibliography}
\bibliographystyle{icml2025}

\newpage
\appendix
\onecolumn

\section{Proof of Theorem \ref{theorem:ours}} \label{proof:ours}

\begin{proof}
The first optimization problem optimizes \( I(x, z) \), which imposes a looser constraint on \( I(o, z) \), as it does not directly regulate the information flow from \( o \) to \( z \). In contrast, the second optimization problem directly constrains \( I(o, z) \), which may result in a smaller \( I(o, z; \phi_o^\star) \). Therefore, we have:
\begin{equation}
    I(o, z; \phi_o^\varepsilon) \geq I(o, z; \phi_o^\star).
\end{equation}

From the optimization objectives of the two problems, it follows that:
\begin{equation}
    I(o, z; \phi_o^\varepsilon) - \frac{1}{\beta} J^\varepsilon \geq I(o, z; \phi_o^\star) - \frac{1}{\beta} J^\star.
\end{equation}

Rearranging this inequality gives:
\begin{equation}
    |J^\varepsilon - J^\star| \leq \beta \cdot \left( I(o, z; \phi_o^\varepsilon) - I(o, z; \phi_o^\star) \right).
\end{equation}

According to the assumption that the mutual information gap is bounded:
\begin{equation}
    I(o, z; \phi_o^\varepsilon) - I(o, z; \phi_o^\star) \leq \frac{\delta}{\beta},
\end{equation}
We substitute this bound into the inequality:
\begin{equation}
    |J^\varepsilon - J^\star| \leq \beta \cdot \frac{\delta}{\beta} = \delta.
\end{equation}

Thus, the performance gap is bounded as:
\begin{equation}
    |J^\star - J^\varepsilon| \leq \delta.
\end{equation}

This completes the proof.
\end{proof}

\section{Details of Experiment Setting}

\subsection{Details of Benchmarks}
\label{sec:appendix_benchmarks}

\subsubsection{CortexBench}

We provide a detailed overview of the four imitation learning benchmarks used in CortexBench~\cite {majumdar2023we}. 
CortexBench is a single-task benchmark that includes 7 selected simulators, collectively offering 17 different embodied AI tasks spanning locomotion, navigation, and both dexterous and mobile manipulation. 
Three of the simulators are primarily designed for reinforcement learning and are therefore excluded from our analysis.
The remaining four simulators, with a total of 14 tasks, are retained for validation: Adroit (2 tasks)~\cite{rajeswaran2018learning}, Meta-World (5 tasks)~\cite{yu2020meta}, DMControl (5 tasks)~\cite{tassa2018deepmind}, and TriFinger (2 tasks)~\cite{wuthrich2021trifinger}.

First, Adroit~\cite{rajeswaran2018learning} is a suite of dexterous manipulation tasks in which an agent controls a 28-DoF anthropomorphic hand. It includes two of the most challenging tasks: Relocate and Reorient-Pen. In these tasks, the agent must manipulate an object to achieve a specified goal position and orientation. Each task consists of 100 demonstrations.

Second, MetaWorld~\cite{yu2020meta} is a collection of tasks in which agents command a Sawyer robot arm to manipulate objects in a tabletop environment. CortexBench includes five tasks from MetaWorld: Assembly, Bin-Picking, Button-Press, Drawer-Open, and Hammer. Each task consists of 25 demonstrations.

Third, DeepMind Control (DMControl)~\cite{tassa2018deepmind} is a widely studied image-based continuous control benchmark, where agents perform locomotion and object manipulation tasks. CortexBench includes five DMC tasks: Finger-Spin, Reacher-Hard, Cheetah-Run, Walker-Stand, and Walker-Walk. Each task consists of 100 demonstrations.

Lastly, TriFinger (TF)~\cite{wuthrich2021trifinger} is a robot consisting of a three-finger hand with 3-DoF per finger. CortexBench includes two tasks from TriFinger: Push-Cube and Reach-Cube. Each task consists of 100 demonstrations.

\begin{figure*}[ht]
\begin{center}
\centerline{\includegraphics[width=\textwidth]{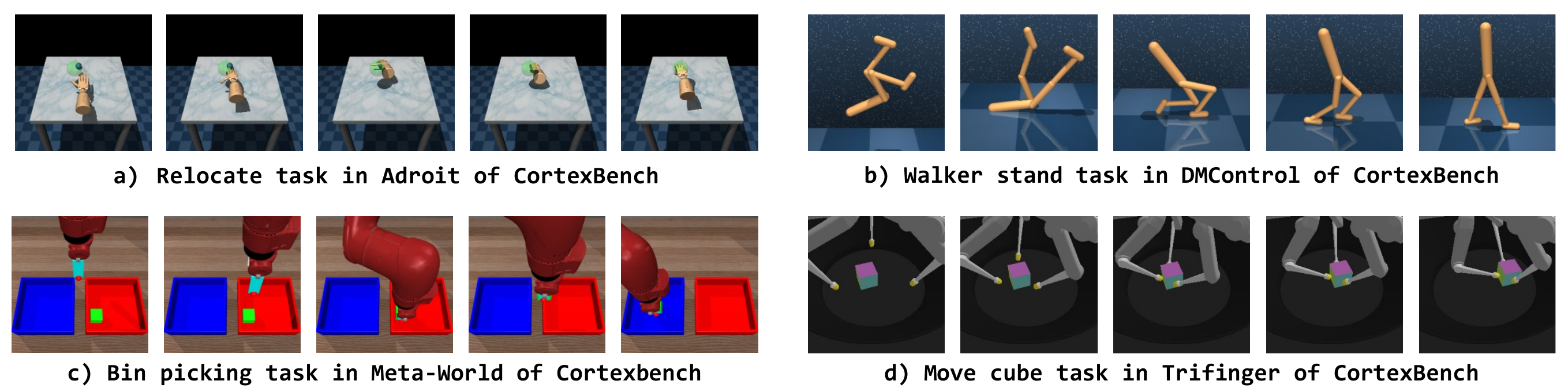}}
\caption{Visualizations for one task from each suite in CortexBench.
}
\label{fig:cortex_case}
\end{center}
\vskip -0.25in
\end{figure*}

Although only Meta-World is strictly a robot manipulation benchmark, we include all tasks to demonstrate the effectiveness of IB comprehensively. 
We provide visualizations for one task from each benchmark, as shown in~\cref{fig:cortex_case}.

\begin{figure*}[ht]
\begin{center}
\centerline{\includegraphics[width=\textwidth]{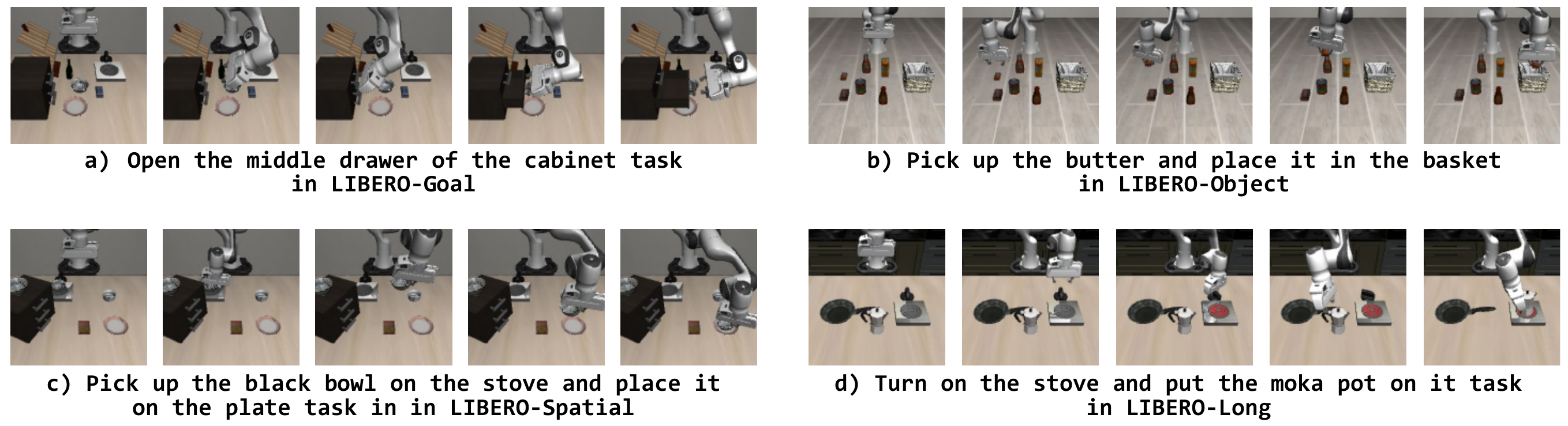}}
\caption{Visualizations for one task from each suite in LIBERO.
}
\label{fig:libero_case}
\end{center}
\vskip -0.25in
\end{figure*}

\subsubsection{LIEBRO}
LIBERO is a language-conditioned multi-task benchmark comprising 130 tasks across five suites. LIBERO~\cite{liu2024libero} has four task suites: LIBERO-Goal (10 tasks), LIBERO-Object (10 tasks), LIBERO-Spatial (10 tasks), and LIBERO-100 (100 tasks).

LIBERO-Goal tasks share the same objects with fixed spatial relationships but differ in task goals, requiring the robot to continually acquire new knowledge about motions and behaviors. 
Examples include (1) opening the middle drawer of the cabinet, (2) opening the top drawer and placing the bowl inside, (3) pushing the plate to the front of the stove, (4) placing the bowl on the plate, (5) placing the bowl on the stove, (6) placing the bowl on top of the cabinet, (7) placing the cream cheese in the bowl, (8) placing the wine bottle on the rack, (9) placing the wine bottle on top of the cabinet, and (10) turning on the stove.

LIBERO-Object tasks involve the robot picking and placing unique objects, requiring it to continually learn and memorize new object types. 
Examples include (1) picking up the alphabet soup and placing it in the basket, (2) picking up the BBQ sauce and placing it in the basket, (3) picking up the butter and placing it in the basket, (4) picking up the chocolate pudding and placing it in the basket, (5) picking up the cream cheese and placing it in the basket, (6) picking up the ketchup and placing it in the basket, (7) picking up the milk and placing it in the basket, (8) picking up the orange juice and placing it in the basket, (9) picking up the salad dressing and placing it in the basket, and (10) picking up the tomato sauce and placing it in the basket.

LIBERO-Spatial requires the robot to place a bowl, selected from the same set of objects, onto a plate. The robot must continually learn and memorize new spatial relationships.
Examples include (1) picking up the black bowl between the plate and the ramekin and placing it on the plate, (2) picking up the black bowl from the table center and placing it on the plate, (3) picking up the black bowl in the top drawer of the wooden cabinet and placing it on the plate, (4) picking up the black bowl next to the cookie box and placing it on the plate, (5) picking up the black bowl next to the plate and placing it on the plate, (6) picking up the black bowl next to the ramekin and placing it on the plate, (7) picking up the black bowl on the cookie box and placing it on the plate, (8) picking up the black bowl on the ramekin and placing it on the plate, (9) picking up the black bowl on the stove and placing it on the plate, and (10) picking up the black bowl on the wooden cabinet and placing it on the plate.

LIBERO-100 consists of 100 tasks involving diverse object interactions and versatile motor skills. It can be divided into LIBERO-10 (10 tasks) and LIBERO-90 (90 tasks), where we use LIBERO-10, also referred to as LIBERO-Long, as our benchmark. 
LIBERO-Long requires the robot to learn long-horizon tasks, demanding it to plan and execute actions over extended periods to accomplish complex objectives.
Examples include (1) turning on the stove and placing the moka pot on it, (2) putting the black bowl in the bottom drawer of the cabinet and closing it, (3) putting the yellow and white mug in the microwave and closing it, (4) putting both moka pots on the stove, (5) putting both the alphabet soup and the cream cheese box in the basket, (6) putting both the alphabet soup and the tomato sauce in the basket, (7) putting the cream cheese box and the butter in the basket, (8) putting the white mug on the left plate and the yellow and white mug on the right plate, (9) putting the white mug on the plate and the chocolate pudding to the right of the plate, and (10) picking up the book and placing it in the back compartment of the caddy.

We provide visualizations for one task from each suite, as shown in~\cref{fig:libero_case}.

\subsection{Details of Baselines}
\label{sec:appendix_baselines}

\subsubsection{Baselines in CortexBench}

In CortexBench, the classification of baselines is primarily based on the visual encoder used.

For full fine-tuning baselines, ResNet~\cite{he2016deep} and ViT~\cite{dosovitskiy2020image} are baselines built from the original ResNet-18 and ViT-S models, using only a portion of their architecture and with uninitialized parameters.

For partial fine-tuning baselines, 
R3M~\cite{nair2023r3m} pre-trains a ResNet model on human videos~\cite{grauman2022ego4d} using time contrastive learning and video-language alignment. For direct comparison, we use the version reproduced with ViT.
VC-1~\cite{majumdar2023we} pre-trains a ViT using
Masked Auto-Encoding (MAE)~\cite{he2022masked} on a mix of human-object interaction videos, navigation, and the ImageNet~\cite{deng2009imagenet} datasets.
Voltron~\cite{karamcheti2023language}, a framework for language-driven representation learning from human videos and associated captions, pre-trains a ViT using MAE.
MPI~\cite{zeng2024learning}, a framework for interaction-oriented representation learning, directs the model to predict transition frames and detect manipulated objects using keyframes as input. It learns from human videos and associated captions.

If a proprioceptive state is available, it is first transformed into embeddings using a linear layer. Depending on the fusion method, these embeddings are then combined with the visual embeddings. For spatial fusion, an MLP is used, while for temporal fusion, a temporal transformer is employed. The fused features are ultimately processed through an MLP-based policy head to generate actions.

\subsubsection{Baselines in LEBERO}

Similar to previous work~\cite{zhu2024spa}, the baselines in LIBERO largely follow the three architectures outlined in the original paper~\cite{liu2024libero}, which we have renamed as BC-RNN, BC-Transformer, and BC-VILT. These three baselines are part of the temporal fusion methods.

BC-RNN uses a ResNet as the visual backbone to encode per-step visual observations, with an LSTM as the temporal backbone to process a sequence of encoded visual information. The language instruction is incorporated into the ResNet features using the FiLM method~\cite{perez2018film}, and is added to the LSTM inputs.

BC-Transformer employs a similar ResNet-based visual backbone but instead uses a transformer decoder~\cite{vaswani2017attention} as the temporal backbone to process outputs from ResNet, which are temporal sequences of visual tokens. The language embedding is treated as a separate token alongside the visual tokens in the input to the transformer.

BC-VILT utilizes a ViT as the visual backbone and a transformer decoder as the temporal backbone. The language embedding is treated as a separate token in the inputs of both the ViT and the transformer decoder. All temporal backbones output a latent vector at each decision-making step.

Additionally, we introduce a spatial fusion method, BC-MLP, which uses a similar ResNet-based visual backbone. The visual and language embeddings are directly concatenated and input into an MLP for fusion. After feature fusion, all methods use an MLP-based policy head to generate actions.

\begin{table*}[ht]
\begin{minipage}[c]{0.48\textwidth}
\makeatletter\def\@captype{table}
\centering
    \caption{Training hyperparameters of all (full $|$ partial) fine-tuning baselines in CortexBench.}
    \vskip -0.05in
    \begin{tabular}{@{}cc@{}}
        \toprule
        Hyperparameters & Training \\
        \midrule
        epoch & 50 $|$ 100  \\
        batch size &  256 $|$ 512  \\
        optimizer & AdamW  \\
        learning rate & 1e-4 $|$ 1e-3  \\
        weight decay & 1e-4 \\
        lr scheduler & Cosine  \\
        lr warm up & 0 \\
        clip grad & 100 \\
        augmentation & Resize, CenterCrop, Normalize \\
        history length & 3 \\
        \bottomrule
    \end{tabular}
    \label{tab:appendix_cortex_para}
\end{minipage}
\begin{minipage}[c]{0.48\textwidth}
\makeatletter\def\@captype{table}
\centering
  \caption{Training hyperparameters of all baselines in LIBERO.}
  \vskip -0.05in
  \begin{tabular}{@{}cc@{}}
    \toprule
    Hyperparameters & Training  \\
    \midrule
    epoch & 50  \\
    batch size &  64  \\
    optimizer &  AdamW  \\
    learning rate & 1e-4  \\
    weight decay & 1e-4 \\
    lr scheduler & Cosine  \\
    lr warm up & 0 \\
    clip grad & 100 \\
    augmentation & Normalize, ColorJitter \\
    history length & 10 \\
    \bottomrule
  \end{tabular}
  \label{tab:appendix_libero_para}
\end{minipage}
\end{table*}

\begin{table*}[ht]
\centering
\caption{IB-related Hyperparameters of all baselines in both CortexBench and LIBERO.
}
\vskip -0.05in
\begin{tabular}{cc}
\toprule
IB-related Hyperparameters & Training \\
\toprule
\multicolumn{2}{c}{\textit{MINE model}} \\
\midrule
architecture & 4-layer MLP \\
hidden size & 512 \\
output size & 1 \\
optimizer & Adam \\
learning rate & 1e-5 \\
loss weight & 0.1 \\
\midrule
\multicolumn{2}{c}{\textit{IB loss}} \\
\midrule
Lagrange multiplier $\beta$ & [1e-4, 1e-2] \\
\bottomrule
\end{tabular}
\label{tab:appendix_ib_para}
\end{table*}

\subsection{Details of Implementations}
\label{sec:appendix_implementations}

For Cortexbench and LIBERO experiments, we use a single NVIDIA V100 or A100 GPU (CUDA 11.8) with 12 CPUs.
For real-world experiments, the single-task setting uses one V100 GPU with 12 CPUs, while the language-conditioned multi-task setting is trained on 8 A100 GPUs with 100 CPUs and evaluated on a single A100 GPU.

\subsubsection{CortexBench}
We largely adhere to the original parameter settings from the CortexBench paper~\cite{majumdar2023we}. 
For both full fine-tuning and partial fine-tuning methods, as shown in~\cref{tab:appendix_cortex_para}, training parameters are presented with full fine-tuning on the left and partial fine-tuning on the right.
For model architecture parameters, spatial fusion employs a 4-layer MLP, where the input dimension matches the output dimension of the image encoder. The features are first downsampled and then upsampled to maintain consistency with the input dimension.
For temporal fusion, each modality's feature dimension is first projected to 64, then processed through a four-layer, six-head Transformer.
For dataset configurations, we adopt a full-shot setting, training with 100 demonstrations for Adroit, 100 for DMControl, 25 for MetaWorld, and 100 for TriFinger. During evaluation, we assess performance using 25, 10, 25, and 25 test trajectories, respectively.

\begin{figure*}[ht]
\begin{center}
\centerline{\includegraphics[width=\textwidth]{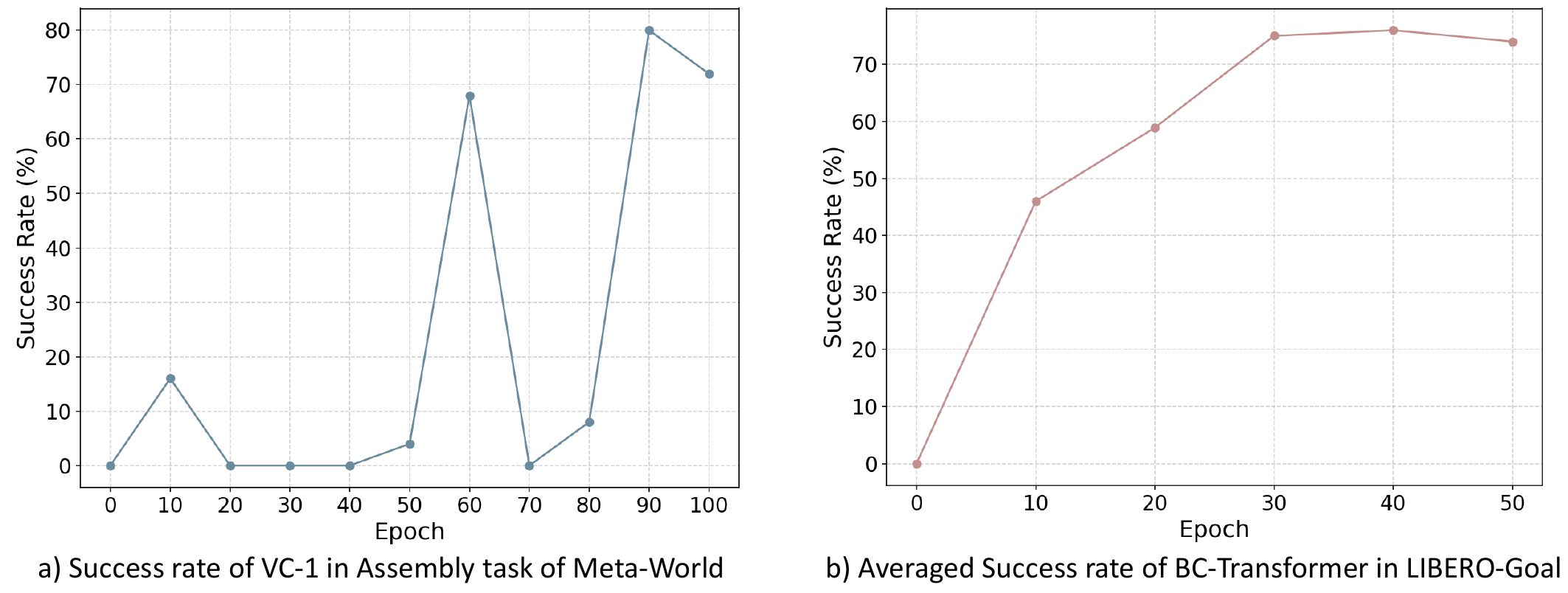}}
\caption{Comparison of success rate curves between single-task and multi-task training.
}
\label{fig:model_selection}
\end{center}
\vskip -0.25in
\end{figure*}

\subsubsection{LIBERO}
We largely follow the original parameter settings from the LIBERO paper~\cite{liu2024libero}. The training parameters are provided in~\cref{tab:appendix_libero_para}.
Regarding model architecture, Appendix A.1 of the original LIBERO paper~\cite{liu2024libero} describes the model parameters for BC-RNN, BC-Transformer, and BC-VILT. Here, we present the model parameters for BC-MLP, which shares the same architecture as BC-Transformer except for the fusion module. Specifically, BC-MLP employs a four-layer MLP with a hidden size of 256 as its fusion module.
For dataset configurations, in the full-shot setting, we use five demonstrations for evaluation, leaving the remaining 45 for training, which is considered the full-shot setting in our experiments. However, full-shot training typically refers to utilizing all 50 demonstrations without allocating any for evaluation, as robotic systems can operate without separate validation data.

\textit{For BC+IB, all training and model parameters remain identical to those of BC, except for the IB-specific parameters. 
}

\subsection{Details of Model Selection}
\label{sec:appendix_model_selection}

\subsubsection{CortexBench}
In the single-task dataset CortexBench, we observed that the learning curves of certain tasks exhibit significant oscillations, such as the assemble task in MetaWorld, as shown in~\cref{fig:model_selection} (a). Previous studies often record performance at intervals of many epochs or steps, selecting either the highest value~\cite{majumdar2023we} or the average of multiple peak values~\cite{ze20243d}. Following~\cite{majumdar2023we}, we directly use the highest value to explore the model's full potential on the given task.

\subsubsection{LIBERO}
For the multi-task dataset LIBERO, we observed that while the learning curve for individual tasks may still oscillate, a decrease in success rate for one task is often accompanied by an increase for another. This trade-off results in smoother overall learning curves across multiple tasks, as shown in~\cref{fig:model_selection} (b). To make model selection more practical and representative, we directly select the model from the final epoch.

\begin{table*}[htbp]
\small
\centering
\caption{Task-wise Performance on Adroit and Trifinger of CortexBench. We evaluated 4 tasks of 2 benchmarks using 3 random seeds and reported the average success rate (sr). 
The best performance is bolded.
}
\begin{tabular}{
>{\raggedright\arraybackslash}m{1.5cm}
>{\centering\arraybackslash}m{2cm}
>{\centering\arraybackslash}m{1.5cm}
>{\centering\arraybackslash}m{2cm}|
>{\centering\arraybackslash}m{2cm}
>{\centering\arraybackslash}m{2cm}
>{\centering\arraybackslash}m{1.5cm}
}
\toprule
           & \multicolumn{3}{c|}{Adroit}                       & \multicolumn{3}{c}{TriFinger}                    \\
\multicolumn{1}{l}{Method} & Reorient-Pen & Relocate & Avg & Reach-Cube & Move-Cube & Avg \\
\midrule
ResNet                                         & 65.33                            & 66.67                        & 66.00                   & 87.12                          & 56.06                         & 71.59                   \\
\multirow{1}{*}{ResNet+IB}                     & \textbf{69.33}                   & \textbf{74.67}               & \textbf{72.00}          & \textbf{87.14}                          & \textbf{57.45}                         & \textbf{72.30}          \\
\midrule
ViT                                            & 61.33                            & 9.33                         & 35.33                   & \textbf{78.77}                 & 32.37                         & 55.57                   \\
\multirow{1}{*}{ViT+IB}                        & \textbf{64.00}                   & \textbf{10.67}               & \textbf{37.33}          & 77.83                          & \textbf{34.04}                & \textbf{55.93}          \\
\midrule
R3M                                            & 45.33                            & \textbf{5.33}                & 25.33                   & 74.29                          & 45.45                         & 59.87                   \\
\multirow{1}{*}{R3M+IB}                        & \textbf{52.00}                   & 2.67                         & \textbf{27.33}          & \textbf{75.04}                 & \textbf{46.23}                & \textbf{60.63}          \\
\midrule
Voltron                                        & 32.00                            & \textbf{5.33}                & 18.67                   & 86.37                          & 62.04                         & 74.21                   \\
\multirow{1}{*}{Voltron+IB}                    & \textbf{38.67}                   & 4.00                         & \textbf{21.33}          & \textbf{86.62}                 & \textbf{63.61}                & \textbf{75.12}          \\
\midrule
VC-1                                           & \textbf{38.67}                   & 10.67                        & 24.67                   & 84.19                          & 59.90                         & 72.05                   \\
\multirow{1}{*}{VC-1+IB}                       & 37.33                            & \textbf{14.67}               & \textbf{26.00}          & \textbf{84.69}                 & \textbf{62.91}                & \textbf{73.80}          \\
\midrule
MPI                                            & 60.00                            & 9.33                         & 34.67                   & 79.69                          & 44.13                         & 61.91                   \\
\multirow{1}{*}{MPI+IB}                        & \textbf{61.33}                   & \textbf{12.00}               & \textbf{36.67}          & \textbf{79.91}                 & \textbf{46.78}                & \textbf{63.34}          \\
\bottomrule
\end{tabular}
\label{tab:cortex_ad_tri_appendix}
\vskip -0.2in
\end{table*}

\begin{table*}[htbp]
\small
\centering
\caption{Task-wise Performance on DMControl of CortexBench. We evaluated 5 tasks using 3 random seeds and reported the average success rate (sr). 
The best performance is highlighted in bold.
}
\begin{tabular}{
>{\raggedright\arraybackslash}m{1.5cm}
>{\centering\arraybackslash}m{2cm}
>{\centering\arraybackslash}m{2cm}
>{\centering\arraybackslash}m{2cm}
>{\centering\arraybackslash}m{2cm}
>{\centering\arraybackslash}m{2cm}
>{\centering\arraybackslash}m{1.5cm}
}
\toprule
Method             & Cheetah-Run    & Finger-Spin    & Reacher-Easy   & Walker-Stand   & Walker-Walk    & Avg            \\
\midrule
ResNet                         & 38.32          & 88.37          & 92.20          & 91.42          & 64.34          & 74.93          \\
\multirow{1}{*}{ResNet+IB}     & \textbf{50.75} & \textbf{90.42} & \textbf{99.78} & \textbf{96.39} & \textbf{87.37} & \textbf{84.94} \\
\midrule
ViT                            & \textbf{7.22}  & 3.39  & 18.97 & \textbf{18.05} & 4.43           & 10.41          \\
\multirow{1}{*}{ViT+IB}        & 4.27  & \textbf{12.34} & \textbf{24.11} & 17.45 & \textbf{4.46}  & \textbf{12.53} \\
\midrule                           
R3M                            & \textbf{17.01} & 65.31 & 53.73          & \textbf{49.25} & 16.27          & 40.31          \\
\multirow{1}{*}{R3M+IB}        & 16.19          & \textbf{72.10} & \textbf{57.34} & 46.48 & \textbf{16.60} & \textbf{41.74} \\
\midrule                           
Voltron                        & 1.65  & 8.56           & \textbf{44.06} & 46.27          & 26.19          & 25.35          \\
\multirow{1}{*}{Voltron+IB}    & \textbf{6.93}  & \textbf{23.85} & 35.89          & \textbf{56.48} & \textbf{42.68} & \textbf{33.16} \\
\midrule                           
VC-1                           & 20.02          & \textbf{85.35} & 74.01          & 64.65          & 25.07          & 53.82          \\
\multirow{1}{*}{VC-1+IB}       & \textbf{21.52} & 80.91 & \textbf{74.80} & \textbf{67.10} & \textbf{30.34} & \textbf{54.93} \\
\midrule                           
MPI                            & \textbf{38.76} & \textbf{88.43} & 75.87          & 68.92          & 25.27          & 59.45          \\
\multirow{1}{*}{MPI+IB}        & 33.82 & 86.44 & \textbf{86.99} & \textbf{69.29} & \textbf{30.50} & \textbf{61.41} \\
\bottomrule
\end{tabular}
\label{tab:cortex_dmc_appendix}
\vskip -0.2in
\end{table*}

\begin{table*}[htbp]
\small
\centering
\caption{Task-wise Performance on Meta-World of CortexBench. We evaluated 5 tasks using 3 random seeds and reported the average success rate (sr). 
The best performance is highlighted in bold.
}
\begin{tabular}{
>{\raggedright\arraybackslash}m{1.5cm}
>{\centering\arraybackslash}m{2cm}
>{\centering\arraybackslash}m{2cm}
>{\centering\arraybackslash}m{2cm}
>{\centering\arraybackslash}m{2cm}
>{\centering\arraybackslash}m{2cm}
>{\centering\arraybackslash}m{1.5cm}
}
\toprule
Method                         & Assembly                 & Bin-Picking              & Button-Press             & Drawer-Open              & Hammer                   & Avg                   \\
\midrule
ResNet                                      & 40.00                    & 74.67                    & 94.67                    & 100.00                   & 96.00                    & 81.07                 \\
\multirow{1}{*}{ResNet+IB}                                   & \textbf{49.33}           & \textbf{76.00}           & \textbf{94.67}           & \textbf{100.00}          & \textbf{96.00}           & \textbf{83.20}        \\
\midrule
ViT                                         & 13.33                    & \textbf{13.33}           & \textbf{21.33}           & 37.33                    & 73.33                    & 31.73                 \\
\multirow{1}{*}{ViT+IB}                                    & \textbf{13.33}           & 9.33                     & 18.67                    & \textbf{62.67}           & \textbf{76.00}           & \textbf{36.00}        \\
\midrule
R3M                                         & \textbf{42.67}           & \textbf{56.00}           & 38.67                    & 66.67                    & 61.33                    & 53.07                 \\
\multirow{1}{*}{R3M+IB}                                      & 38.67                    & 53.33                    & \textbf{38.67}           & \textbf{68.00}           & \textbf{72.00}           & \textbf{54.13}        \\
\midrule
Voltron                                     & \textbf{60.00}           & 58.67                    & \textbf{68.00}           & 82.67                    & 93.33                    & 72.53                 \\
\multirow{1}{*}{Voltron+IB}                                  & 57.33                    & \textbf{74.67}           & 54.67                    & \textbf{93.33}           & \textbf{92.00}           & \textbf{74.40}        \\
\midrule
VC-1                                        & 68.00                    & 60.00                    & 65.33                    & 100.00                   & 94.67                    & 77.60                 \\
\multirow{1}{*}{VC-1+IB}                                     & \textbf{70.67}           & \textbf{76.00}           & \textbf{69.33}           & \textbf{100.00}          & \textbf{96.00}           & \textbf{82.40}        \\
\midrule
MPI                                         & 61.33                    & 40.00                    & 58.67                    & 100.00                   & 72.00                    & 66.40                 \\
\multirow{1}{*}{MPI+IB}                                      & \textbf{61.33}           & \textbf{53.33}           & \textbf{58.67}           & \textbf{100.00}          & \textbf{73.33}           & \textbf{69.33}        \\
\bottomrule
\end{tabular}
\label{tab:cortex_mw_appendix}
\vskip -0.2in
\end{table*}

\section{Additional Experiment Results}

\subsection{Details of Simulation Experiments}
\label{sec:appendix_task-wise_exp}

We provide the task-wise results and corresponding 
$\beta$ values for CortexBench. The results for Adroit and TriFinger in~\cref{tab:cortex_ad_tri_appendix}, DMControl in~\cref{tab:cortex_dmc_appendix}, and MetaWorld are shown in~\cref{tab:cortex_mw_appendix}.
Across almost all tasks in CortexBench, incorporating the IB consistently improves performance compared to vanilla BC methods. Notably, models such as ResNet+IB, VC-1+IB, and MPI+IB often achieve the highest success rates, demonstrating the benefits of redundancy reduction in latent representations. In most cases, properly tuning $\beta$ (e.g., selecting values in the range of 1e-4 to 1e-2) leads to noticeable improvements.

\begin{figure*}[ht]
\begin{center}
\centerline{\includegraphics[width=0.8\textwidth]{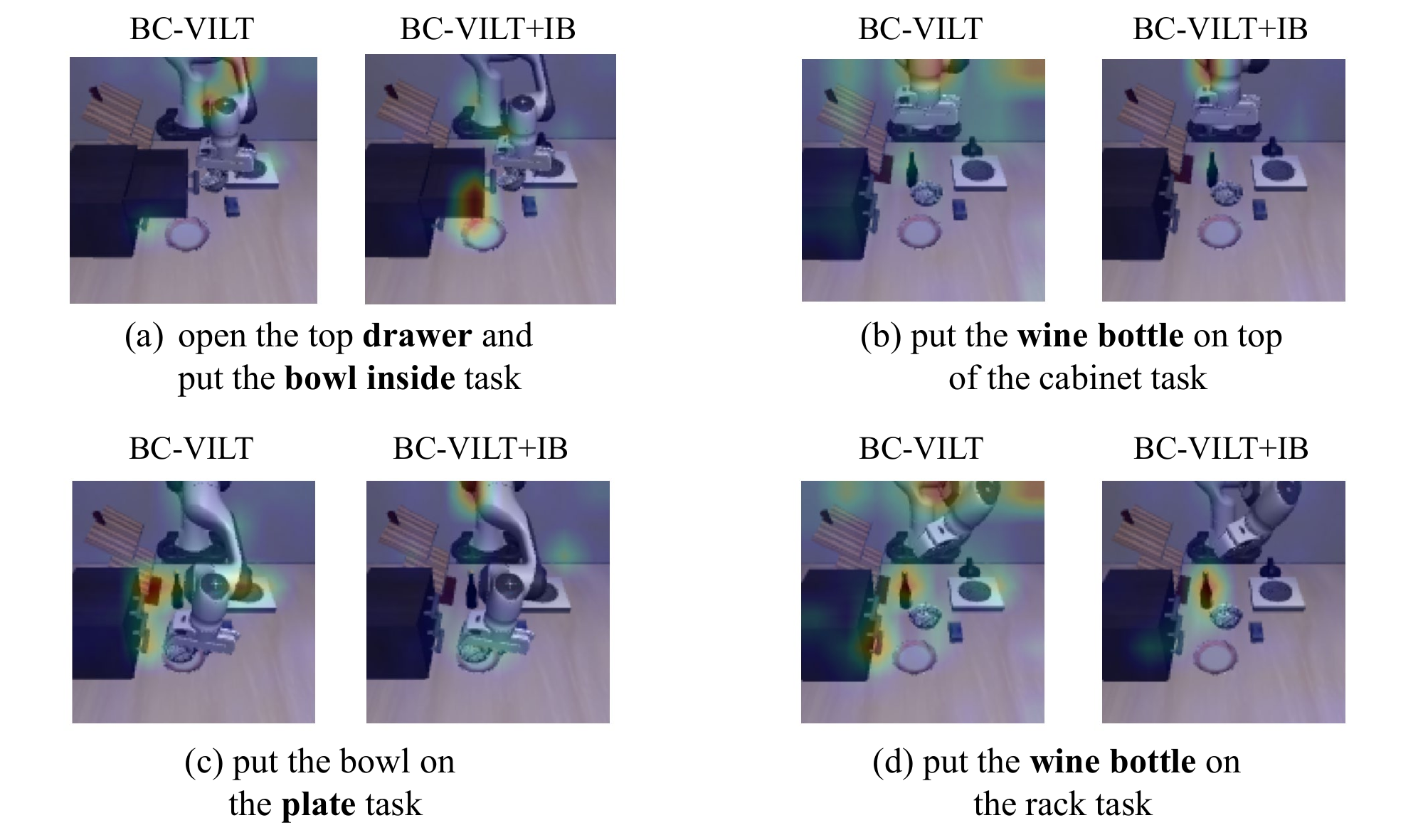}}
\caption{ IB encourages the model to focus more selectively on task-relevant regions, specifically the robotic arm and the target object, by suppressing attention to redundant background features.
}
\label{fig:attn}
\end{center}
\vskip -0.25in
\end{figure*}

\subsection{Attention Map Visualizations}

As shown in~\cref{fig:attn}, IB helps the model focus on task-relevant regions (e.g., the arm and target object) by suppressing attention to redundant background features—an effect less evident without redundancy reduction, highlighting IB’s distinct role beyond standard regularization.

\subsection{More Explanation on the LIBERO Experiments}

\subsubsection{More Experiments on LIBERO-Long}
\label{subsub: appendix_libreo-long}

In the main text, we point out that the experimental results on LIBERO-Long are affected by the limited capacity of baseline models (e.g., BC-VILT with 10M parameters), leading to only marginal improvements. While these models perform well on simpler tasks such as LIBERO-Goal, achieving an 80\% success rate with an 8\% improvement, they show limited gains on more complex tasks like LIBERO-Long, with only a 2\% increase in performance. This is primarily due to the performance ceiling imposed by the lightweight baselines. To further illustrate this limitation, we conduct experiments on LIBERO-Long using Diffusion Policy (DP), where the 1.14M-parameter MLP head of BC-Transformer is replaced with the more expressive 90M-parameter DP head. 
We train DP from scratch on a single A100 GPU with a batch size of 64, a learning rate of 1e-4, and for 50 epochs. The training setup for DP+IB is identical to that of DP, except for the IB-related components. For the IB-specific hyperparameters, the Lagrange multiplier $\beta$ is set to 1e-5.

\begin{table}[htbp]
\centering
\caption{Performance on LIBERO-Long.}
\label{tab:libero_long}
\begin{tabular}{cc}
\toprule
Method & LIBERO-Long \\
\midrule
DP & 78.0 \\
DP+IB & \textbf{84.0} \\
\bottomrule
\end{tabular}
\end{table}

Attention: Our experimental setup differs from that of papers~\cite{kim2024openvla, wen2024diffusion, wen2025dexvla} like OpenVLA. In those works, the image observations are saved at a resolution of 256×256 (instead of 128×128) and undergo additional filtering, such as removing "no-op" (zero) actions and unsuccessful demonstrations. In contrast, our setting uses the raw LIBERO data with lower-resolution images and no filtering.

\subsubsection{Performance on LIBERO-Object}
On LIBERO-Object, we observe that the success rate does not consistently improve with an increasing number of demonstrations. Specifically, in the 10-shot setting, BC-VILT achieves a success rate of 56.17\%, whereas in the full-shot setting, its performance drops to 43.00\%. We hypothesize that this decline is due to inherent data distribution characteristics and suboptimal data quality within the benchmark.

\begin{wrapfigure}[16]{r}{0.5\textwidth}
  \vspace{-40pt}
  \begin{center}
    \includegraphics[width=0.48\textwidth]{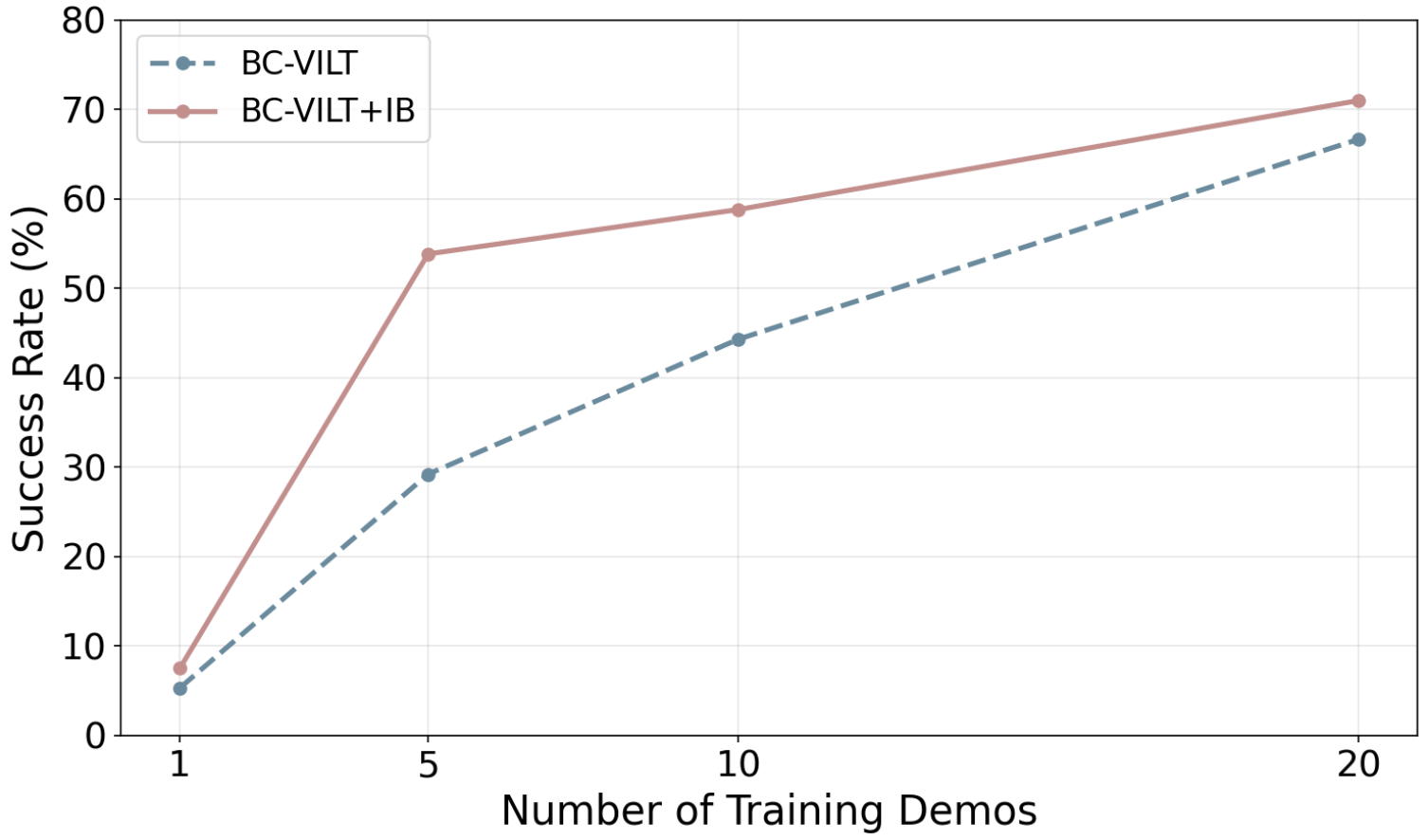}
  \end{center}
  \vspace{-10pt}
  \caption{Comparison of the success rates of BC-VILT+IB trained with 1, 5, 10, and 20 demonstrations against the vanilla BC-VILT in the LIBERO-Goal suite.}
  \label{fig:appendix_few-shot}
\end{wrapfigure}

\subsection{Extension to Few-shot Setting}
We further assess the effectiveness of IB in few-shot settings by evaluating BC-VILT under varying numbers of demonstrations in LIBERO-Goal, as shown in~\cref{fig:appendix_few-shot}.
The results consistently show that incorporating IB improves success rates across all LIBERO suites, demonstrating its efficacy in few-shot learning scenarios.


\end{document}